\documentclass{article}

\PassOptionsToPackage{numbers, compress}{natbib}

\usepackage[preprint]{neurips_2025}

\usepackage[utf8]{inputenc} 
\usepackage[T1]{fontenc}    
\usepackage{hyperref}       
\usepackage{url}            
\usepackage{booktabs}       
\usepackage{amsfonts}       
\usepackage{nicefrac}       
\usepackage{microtype}      
\usepackage{xcolor}         

\usepackage{amsmath}
\usepackage{mathtools}
\usepackage{algorithm}
\usepackage{algorithmic}
\usepackage{multirow}
\usepackage{colortbl}
\usepackage{xcolor}
\usepackage{graphicx}
\usepackage{amssymb}
\usepackage{amsthm}
\usepackage{wrapfig}
\usepackage[most]{tcolorbox}
\usepackage{makecell}

\definecolor{calmblue}{rgb}{0.7, 0.65, 0.98}  
\definecolor{lighterblue}{rgb}{0.82, 0.78, 0.99} 

\newtheorem{theorem}{Theorem}[section]

\newtheorem{lemma}[theorem]{Lemma}
\newtheorem{corollary}[theorem]{Corollary}
\newtheorem{definition}[theorem]{Definition}

\title{SetPINNs: Set-based Physics-informed \\Neural Networks}

\author{Mayank Nagda \\
RPTU Kaiserslautern-Landau, Germany \\
\texttt{nagda@cs.uni-kl.de}\\
\And
Phil Ostheimer\\
RPTU Kaiserslautern-Landau, Germany \\
\texttt{ostheimer@cs.uni-kl.de}\\
\And
Thomas Specht\\
RPTU Kaiserslautern-Landau, Germany \\
\texttt{thomas.specht@rptu.de}\\
\And
Frank Rhein\\
Karlsruhe Institute of Technology, Germany \\
\texttt{frank.rhein@kit.edu}\\
\And
Fabian Jirasek\\
RPTU Kaiserslautern-Landau, Germany \\
\texttt{fabian.jirasek@rptu.de}\\
\And
Stephan Mandt\\
University of California, Irvine, USA \\
\texttt{mandt@uci.edu}\\
\And
Marius Kloft\\
RPTU Kaiserslautern-Landau, Germany \\
\texttt{kloft@cs.uni-kl.de}\\
\And
Sophie Fellenz\\
RPTU Kaiserslautern-Landau, Germany \\
\texttt{fellenz@cs.uni-kl.de}\\
}

\begin{document}

\maketitle

\begin{abstract}
  Physics-Informed Neural Networks (PINNs) solve partial differential equations using deep learning. However, conventional PINNs perform pointwise predictions that neglect dependencies within a domain, which may result in suboptimal solutions. We introduce SetPINNs, a framework that effectively captures local dependencies. With a finite element-inspired sampling scheme, we partition the domain into sets to model local dependencies while simultaneously enforcing physical laws. We provide a rigorous theoretical analysis showing that SetPINNs yield unbiased, lower-variance estimates of residual energy and its gradients, ensuring improved domain coverage and reduced residual error. Extensive experiments on synthetic and real-world tasks show improved accuracy, efficiency, and robustness.
\end{abstract}

\section{Introduction}
\label{introduction}

Modeling complex physical systems by solving partial differential equations (PDEs) is a central challenge in many scientific and engineering fields. Deep learning approaches, especially PINNs \citep{raissi2019physics}, have emerged as a promising framework for integrating physical laws (PDEs) directly into the learning process. PINNs have been successfully applied to many physical systems, including fluid mechanics \citep{cai2021physics} and climate forecasting \citep{verma2024climode}, among others.

Conventional PINNs and variants employ multilayer perceptron (MLP) and perform pointwise prediction over the domain. Whilst remarkably successful in various scenarios, recent research has shown that PINNs may fail in scenarios where PDE solutions exhibit high-frequency or multiscale features \citep{fuks2020limitations,leiteritz2021avoid,krishnapriyan2021characterizing,mojgani2022lagrangian,daw2022mitigating,wang2022and,zhao2024pinnsformer,wu2024ropinn}. In such cases, the optimization process tends to get stuck in local minima, resulting in overly smooth suboptimal approximations of ground-truth solutions.

To mitigate these \textit{failure modes}, researchers have proposed data interpolation techniques and tailored training strategies \citep{krishnapriyan2021characterizing,wang2021understanding,wang2022and}. More recently, research has begun to focus on exploiting inherent domain dependencies (such as temporal or regional correlations) \citep{wu2024ropinn,zhao2024pinnsformer}. Although these methods show promise, they remain ad hoc, require additional samples, and introduce significant computational costs, thereby limiting their applicability in many real-world contexts.

\begin{figure}[h]
    \centering
    \includegraphics[scale=0.66]{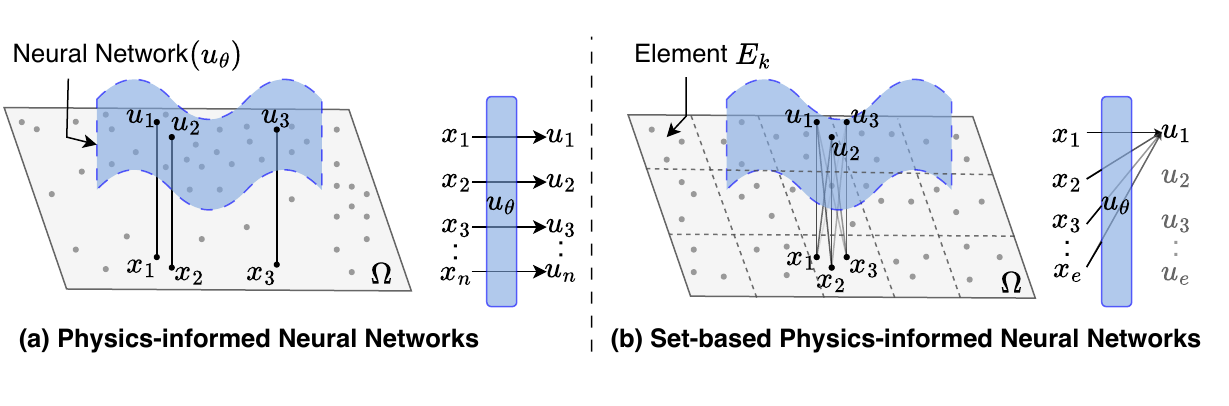}
    \caption{
    Comparison between (a) standard Physics-Informed Neural Networks (PINNs) and (b) the proposed Set-based PINNs (SetPINNs). Standard PINNs predict \( u_\theta(x_i) \) independently at sparse, uniformly sampled points \( x_i \in \Omega \), often underrepresenting parts of the domain \(\Omega\) and lacking local context. SetPINNs partition the domain into local elements \( E_k \), jointly conditioning predictions on sampled sets and capturing local structure. This promotes better coverage, stronger inductive bias, and improved physical consistency.
}
    \label{fig:main-fig-1}
\end{figure}

Orthogonal to existing approaches, we focus on the foundational issue of capturing local dependencies that are often overlooked in existing PINNs. As shown in Figure~\ref{fig:main-fig-1}, conventional PINNs operate at isolated points, making independent predictions at each location. This introduces two key limitations: (1) uniform sampling can leave large regions of the domain unsupervised, and (2) the model lacks mechanisms to exploit correlations among neighboring points. We hypothesize that these shortcomings contribute to instability and poor generalization.

To overcome these challenges, we propose SetPINNs---a new paradigm for learning physics-informed solutions by incorporating domain dependencies in \textit{any} physical domain. Inspired by finite element methods, SetPINNs partition the domain into multiple elements and process them to aggregate local information. Sets are derived from each element using an element-aware sampling technique. These sets are then processed through an attention-based architecture, allowing the framework to capture local dependencies accurately and efficiently. Furthermore, unlike traditional PINNs, SetPINNs are well-suited for permutation-invariant systems, where modeling interactions among neighboring points within the domain is natural, enabling a wide range of applications. In summary, our key contributions are as follows:
\begin{itemize}
\item \textbf{Novel Method.} We propose SetPINNs, a set-based framework that captures local dependencies in physical domains by processing point sets to model intra-domain interactions.

\item \textbf{Theoretical Analysis.} We provide a rigorous theoretical analysis showing that SetPINNs yield unbiased, lower-variance estimates of residual energy and its gradients, theoretically ensuring improved domain coverage and reduced residual error.

\item \textbf{Empirical Validation.} Across diverse PDEs and real-world tasks, SetPINNs reduce errors by up to 74\%, converge faster, and show lower variance than state-of-the-art models, demonstrating superior accuracy, efficiency, and robustness.
\end{itemize}

\section{Related Work}
\label{sec:related_work}

\paragraph{Physics-informed Neural Networks.}
Neural network-based methods for solving PDEs have existed for decades \citep{lagaris1998artificial,psichogios1992hybrid}. However, advances in deep learning have led to a resurgence of this idea through PINNs \citep{raissi2019physics}. PINNs incorporate the PDE residual into the loss function, penalizing predictions that violate the underlying PDE by performing regression on the gradients. The points where the PDEs are enforced are called collocation points. Applications of PINNs are not only limited to white-box scenarios but also extend to grey-box scenarios. Whereas white-box scenarios require no real-world data but only a sample of collocation points, grey-box scenarios combine data and collocation points. Examples include the simulation of cardiovascular systems \citep{raissi2020hidden}, fluid mechanics \citep{cai2021physics}, and climate forecasting \citep{verma2024climode}. Extensions of PINNs include architectural innovations \citep{bu2021quadratic,wong2022learning}, optimization techniques \citep{wang2022and,wu2024ropinn}, regularization methods \cite{krishnapriyan2021characterizing}, and improved sampling strategies \cite{wu2023comprehensive}, leading to significant advancements in their performance and applicability.

\paragraph{Failure Modes of PINNs.}
The pointwise prediction nature of conventional PINNs, typically based on multilayer perceptron (MLP), poses significant challenges for optimization, particularly in complex PDE regimes. \citet{wang_understanding_2021} attribute this difficulty to the stiffness of the gradient arising from multiscale interactions in loss, which imposes stringent learning rate constraints. Other studies have identified similar limitations when addressing PDEs with high-frequency or multiscale features \citep{raissi2018deep,fuks2020limitations,krishnapriyan2021characterizing,wang2022and,leiteritz2021avoid,zhao2024pinnsformer}, where optimization frequently converges to local minima, leading to overly smooth and physically inaccurate solutions.

\paragraph{Mitigating Failure Modes.}
Attempts to improve PINNs span training strategies, data interpolation, architectural changes, and domain-aware modeling. Training strategies like Seq2Seq \citep{krishnapriyan2021characterizing} and others \citep{mao2020physics,wang2021understanding,wang2022and} are computationally costly and often unstable. Data-driven approaches \citep{han2018solving,zhu2019physics,chen2021physics} rely on real or synthetic data, which may be scarce. Architectures such as QRes \citep{bu2021quadratic} and FLS \citep{wong2022learning} improve training but face scalability issues. Domain-aware methods like FBPINNs \citep{moseley2023finite} use fixed basis functions, PINNsFormer \citep{zhao2024pinnsformer} captures temporal but not spatial structure, and RoPINNs \citep{wu2024ropinn} optimize regions without modeling joint interactions. In contrast, SetPINNs offer an orthogonal perspective and learn local dependencies from point sets via attention, offering greater flexibility, improved domain coverage, and better accuracy without manual decomposition.

\paragraph{Deep Learning on Sets.}
Permutation-invariant models like Deep Sets \citep{zaheer2017deep,wagstaff2022universal} and Set-Transformers \citep{lee2019set} enable learning over unordered inputs and have powered advances in chemistry, vision, and particle physics \citep{serviansky2020set2graph,zhang2023molsets}. Yet, despite the ubiquity of set-structured physical data, their integration into physics-informed learning remains limited. SetPINNs close this gap by combining set-based modeling with physical constraints, enabling accurate, flexible, and domain-consistent learning.

\section{Methodology}
\label{sec:method}
This section presents preliminaries in Section~\ref{sec:preliminaries} and the proposed SetPINNs approach in Section~\ref{sec:main-method}.

\subsection{Preliminaries}
\label{sec:preliminaries}
We briefly review PINNs and FEMs, highlighting the core principles and connection to SetPINNs.

\paragraph{Problem Setting and PINNs.}
We consider a differential equation defined over a domain \(\Omega \subset \mathbb{R}^d\), with solution \(u: \mathbb{R}^d \to \mathbb{R}^m\). The domain's interior, boundary, and optionally an initial subset are denoted by \(\Omega\), \(\partial\Omega\), and \(\Omega_0\), respectively. Differential operators \(\mathcal{O}_\Omega\), \(\mathcal{O}_{\Omega_0}\), and \(\mathcal{O}_{\partial\Omega}\) encode the governing equation, initial conditions (ICs), and boundary conditions (BCs). For example, the heat equation is written as \(\mathcal{O}_\Omega(u)(x) = u_t - u_{xx}\). The complete problem formulation is:
\begin{equation}
\label{eq:pde}
\mathcal{O}_\Omega(u)(x) = 0,\ x \in \Omega; \quad \mathcal{O}_{\Omega_0}(u)(x) = 0,\ x \in \Omega_0; \quad \mathcal{O}_{\partial\Omega}(u)(x) = 0,\ x \in \partial\Omega.
\end{equation}

PINNs \citep{raissi2019physics} approximate the solution \(u\) with a neural network \(u_\theta\), trained to minimize the residuals of the governing constraints:
\begin{equation}
\label{eq:pinn_loss}
\mathcal{L}(u_\theta) = \sum_{X \in \{\Omega, \Omega_0, \partial\Omega\}} \frac{\lambda_X}{N_X} \sum_{i=1}^{N_X} \left\| \mathcal{O}_X(u_\theta)(x_X^{(i)}) \right\|^2,
\end{equation}
where \(N_X\) is the number of collocation points in region \(X\), and \(\lambda_X\) weights each term. Additional data-based terms may be added, as discussed in Appendix~\ref{app:pinn-loss}.

\paragraph{Finite Element Methods (FEMs).}
FEMs \citep{zienkiewicz2005finite} solve PDEs by partitioning the domain \(\Omega\) into \(N\) non-overlapping elements (e.g., triangles or quadrilaterals) and approximating the solution as a linear combination of local basis functions: \(u_h(x) = \sum_{i=1}^N U_i \phi_i(x)\), where \(\{\phi_i(x)\}\) are element-wise basis functions and \(U_i\) are coefficients. A weighted residual approach is used to derive a global system of algebraic equations, which is then solved for an approximate solution. FEMs exhibit two important properties: (1) \textit{local dependency}, as neighboring points are coupled through overlapping basis support and (2) \textit{global consistency} through inter-element coupling. By explicitly modeling local interactions, FEMs have shown superior accuracy over PINNs in certain complex scenarios \cite{grossmann2024can}. However, PINNs remain attractive for their flexibility, lower complexity, and ability to incorporate real-world data. With SetPINNs, we combine the flexibility of PINNs with the locality principles of FEMs by extending pointwise predictions to element-based predictions. 

\begin{figure*}
\begin{center}
    \includegraphics[scale=0.8]{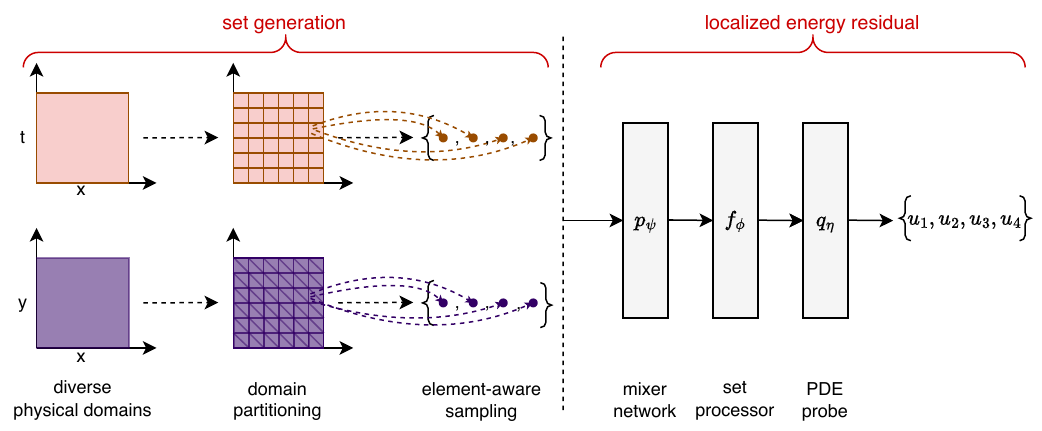}
    \caption{Overview of the SetPINNs framework. On the left, we exemplify the set generation process for diverse physical domains. Sets are formed from neighboring points in the domain. Each set is then processed through a Mixer Network, Set Processor, and PDE Probe to predict the solution for each element in the set. The parameters of SetPINNs are learned using a localized energy residual.}
    \label{fig:main-fig}
\end{center}
\end{figure*}

\subsection{SetPINNs}
\label{sec:main-method}

This section presents the SetPINNs framework. We begin by introducing the element-aware sampling strategy that ensures better domain coverage and forms point sets, then describe the architecture (Figure~\ref{fig:main-fig}) to model local dependencies, and finally derive the set-based physics loss.

\subsubsection{Domain Partitioning and Sampling}
The domain is partitioned into non-overlapping elements, following the standard FEM approach.

\begin{definition}[Domain Partitioning]
\label{def:dp}
\textit{Let \(\Omega \subset \mathbb{R}^d\) be an open, bounded domain. We partition \(\Omega\) into non-overlapping subdomains (elements) \(\{E_k\}_{k=1}^K\) such that
\[
\bigcup_{k=1}^K E_k = \Omega, \quad E_i \cap E_j = \varnothing \text{ for } i \neq j.
\]
We assume that the partition is quasi-uniform, i.e., \( |E_k| \approx h^d \) for some \( h > 0 \).
}
\end{definition}

After partitioning, collocation points are sampled within each element. 

\begin{definition}[Element-Aware Sampling (EAS)]
\label{def:eas}
\textit{Within each element \(E_k\), we uniformly sample \( m_k \) collocation points \(\{x_{k,i}\}_{i=1}^{m_k} \subset E_k\). Let \( M = \sum_{k=1}^K m_k \) denote the total number of collocation points. We require that the sampling density is uniform across elements, i.e., 
\[
\frac{m_k}{|E_k|} = \frac{m_j}{|E_j|} \quad \text{for all } k,j,
\]
which guarantees that each subdomain is neither over- nor under-sampled relative to its size.}
\end{definition}

The element-aware sampling strategy offers two key advantages: it ensures better domain coverage and produces localized point sets. These sets enable the architecture to model local dependencies.

\subsubsection{Architecture}
The SetPINN architecture is designed to capture local dependencies and make joint predictions over multiple points within each element. As shown in Figure~\ref{fig:main-fig}, each element \(E_k\) yields a set \(\mathcal{S}_k\) of collocation points, which is passed through three main components: a Mixer Network, an attention-based Set Processor, and a PDE Probe. We now describe each of these in detail.

\paragraph{Set Generator.}
Given the domain partitioning (Def.~\ref{def:dp}) and element-aware sampling (Def.~\ref{def:eas}), each element \(E_k\) yields a set of collocation points that represent the subdomain \(\Omega_k\). The set generator collects these points into a set \(\mathcal{S}_k\), defined as:
\begin{equation*}
\label{set-generator}
    \Omega_k \xRightarrow{\textmd{set-generator}} \mathcal{S}_k := \{ x^{1}_k, x^{2}_k, \dots, x^{m_k}_k \},
\end{equation*}
where each \(x^i_k \in \Omega_k\). Predicting the PDE solution within \(\Omega_k\) is thus formulated as a set-to-set learning task.

\paragraph{Mixer Network.}
Each point \(x_k^i \in \mathcal{S}_k\) lies in a low-dimensional coordinate space \(\mathbb{R}^d\), which does not adequately capture expressive patterns for PDE approximation. To address this, a pointwise embedding network \(p_\psi\) maps each input into a higher-dimensional representation space \(\mathbb{R}^r\), producing a transformed set \(\mathcal{M}_k = \{ \mathcal{M}_k^i \}_{i=1}^{m_k}\), where \(\mathcal{M}_k^i \in \mathbb{R}^r\). This embedding increases representational capacity and prepares the set for downstream interaction modeling via attention.

\paragraph{Set Processor.}
To model interactions within each set \(\mathcal{M}_k\), we apply a permutation-equivariant attention module \(f_\phi\), implemented using a Transformer encoder block \cite{vaswani2017attention}. For each point \(i\), the processor aggregates contextual information from all other points in the set using:

\[
\mathcal{O}_k^i = \text{MLP}\left( \sum_{j=1}^{m_k} \alpha(x_k^i, x_k^j) \cdot W_V \mathcal{M}_k^j \right), \quad \text{for } i = 1, \dots, m_k,
\]

where \(\alpha(x_k^i, x_k^j)\) are attention weights computed via scaled dot-product similarity between learned embeddings of \(\mathcal{M}_k^i\) and \(\mathcal{M}_k^j\), and \(W_V \in \mathbb{R}^{r \times r}\) is a learnable weight matrix. The MLP is shared across all points, ensuring permutation equivariance.

Unlike sequence-based models, we avoid positional encodings to maintain equivariance under permutations of the input set, an essential property for consistent set-based learning.

\paragraph{PDE Probe.}
The output features \(\mathcal{O}_k = \{ \mathcal{O}_k^i \}_{i=1}^{m_k},\; \mathcal{O}_k^i \in \mathbb{R}^r\) are decoded into the final PDE solution predictions \(u_k = \{ u_k^i \}_{i=1}^{m_k},\; u_k^i \in \mathbb{R}^c\) through a pointwise decoder \(q_\eta\). This module consists of a shallow MLP applied independently to each \(\mathcal{O}_k^i\), mapping the feature space \(\mathbb{R}^r\) to the solution space \(\mathbb{R}^c\). As each prediction is conditioned on a context-aware representation, the model produces locally consistent and structure-informed approximations of the solution within each element.

\subsubsection{Learning Objective}
Conventional PINNs enforce PDE constraints at individual collocation points, treating all locations uniformly. In contrast, SetPINN aggregates residuals within each element, promoting local physical consistency while preserving global balance.

For any constraint region \(X \in \{\Omega,\, \Omega_0,\, \partial\Omega\}\), we define the \emph{localized residual energy} over an element \(E_k\) as:
\begin{equation}
\label{eq:localized_energy}
\mathcal{E}_X(E_k, u_\theta) := \int_{E_k} \bigl\|\mathcal{O}_X(u_\theta)(x)\bigr\|^2 dx \approx \frac{|E_k|}{m_k} \sum_{i=1}^{m_k} \bigl\|\mathcal{O}_X(u_\theta)(x_k^{(i)})\bigr\|^2,
\end{equation}
where the integral is approximated using \(m_k\) collocation points within \(E_k\).

The total SetPINNs loss is then the average localized energy across all elements:
\begin{equation}
\label{eq:setpinns_loss_energy}
\mathcal{L}_{\text{SetPINNs}} = \sum_{X \in \{\Omega,\, \Omega_0,\, \partial\Omega\}} \frac{\lambda_X}{K} \sum_{k=1}^{K} \mathcal{E}_X(E_k, u_\theta),
\end{equation}
where \(\theta = \{\psi, \phi, \eta\}\) denotes the learnable parameters of the Mixer Network \(p_\psi\), Set Processor \(f_\phi\), and PDE Probe \(q_\eta\), which collectively define the function \(u_\theta\) used to predict the solution. Training follows standard PINN protocols and is detailed in Appendix~\ref{app:training_setpinns} for completeness.

\section{Theoretical Analysis}
\label{sec:theory}

We present a theoretical analysis showing that SetPINNs yield statistically improved estimates of PDE residual energy compared to conventional pointwise methods. Specifically, we prove that the Element-Aware Sampling (EAS) strategy produces unbiased, lower-variance estimates of both the residual and its gradients relative to Global Uniform Sampling (GUS). These results imply improved domain coverage and more stable optimization signals, which are essential for effective training in physics-informed models. Formal proofs are provided in the Appendix~\ref{app:theoretical_analysis}. 

We begin by introducing the coverage ratio, a metric that quantifies how well the sampled points capture the residual in the domain. The concept of coverage ratio aligns with residual-based error estimation techniques in numerical methods \cite{quarteroni2008numerical}.

\begin{definition}[Coverage Ratio]
\label{def:coverage_ratio}
\textit{
Let \(u_\theta\) be a neural network with residual \(R_\theta(x) = \mathcal{O}_\Omega(u_\theta)(x)\). The coverage ratio is defined as the ratio \(\frac{I}{\hat{I}}\) of the true residual energy \(I = \int_\Omega |R_\theta(x)|^2\,dx\) to its discrete estimate \(\hat{I}\), where \(\hat{I}_{\mathrm{GUS}} = \frac{|\Omega|}{M} \sum_{j=1}^{M} |R_\theta(x_j)|^2\) and \(\hat{I}_{\mathrm{EAS}} = \sum_{k=1}^K \frac{|E_k|}{m_k} \sum_{i=1}^{m_k} |R_\theta(x_{k,i})|^2\) are the estimates under Global Uniform Sampling (GUS) and Element-Aware Sampling (EAS), respectively.
}
\end{definition}

A coverage ratio close to 1 suggests the sampled points effectively capture the magnitude and distribution of the residual. A much larger value suggests the sampling strategy underestimates the true residual integral, often due to missing regions of high residual.

We first establish that both EAS and GUS provide unbiased estimates of the residual energy.

\begin{lemma}[Unbiased Estimation via Sampling]
\label{lem:unbiased_estimation}
Under uniform random sampling, where points are drawn within each element \(E_k\) for EAS and over the entire domain \(\Omega\) for GUS, the estimators \(\hat{I}_{\mathrm{EAS}}\) and \(\hat{I}_{\mathrm{GUS}}\) (as defined in Definition~\ref{def:coverage_ratio}) are unbiased estimators of the true integral \(\int_\Omega |R_\theta(x)|^2 dx\).
\end{lemma}
\noindent
\textit{Proof Sketch.} Both $\hat{I}_{\mathrm{GUS}}$ and $\hat{I}_{\mathrm{EAS}}$ are scaled sums of samples. For GUS, sampling is from $\mathcal{U}(\Omega)$; for EAS, sampling is from $\mathcal{U}(E_k)$ within each element. By linearity of expectation, both estimators yield $\int_\Omega |R_\theta(x)|^2 dx$ as their expected value. Complete proof is available in Appendix~\ref{app:proof_coverage}.\qed

We now show that EAS yields a statistically more reliable estimator of the residual energy.

\begin{theorem}[Variance Reduction with EAS]
\label{thm:variance_reduction}
Under the same assumptions as in Lemma~\ref{lem:unbiased_estimation}, EAS provides an estimator with equal or lower variance: \(\mathrm{Var}(\hat{I}_{\mathrm{EAS}}) \le \mathrm{Var}(\hat{I}_{\mathrm{GUS}})\).
\end{theorem}
\noindent
\textit{Proof Sketch.} Calculation of variances shows $\mathrm{Var}(\hat{I}_{\mathrm{EAS}}) \le \mathrm{Var}(\hat{I}_{\mathrm{GUS}})$ reduces to proving $\sum_{k=1}^K \frac{1}{|E_k|} \left( \int_{E_k} |R_\theta(x)|^2\, dx \right)^2 \ge \frac{1}{|\Omega|} \left( \int_\Omega |R_\theta(x)|^2\, dx \right)^2$, which is always true in stratified sampling \cite{caflisch1998monte}. Full proof in Appendix~\ref{app:proof_coverage}. \qed

Since both $\hat{I}_{\mathrm{EAS}}$ and $\hat{I}_{\mathrm{GUS}}$ are unbiased estimators of the true total PDE residual energy $I(\theta)$ (Lemma~\ref{lem:unbiased_estimation}), the lower variance of $\hat{I}_{\mathrm{EAS}}$ (Theorem~\ref{thm:variance_reduction}) directly implies a lower Mean Squared Error for the estimator $\hat{I}_{\mathrm{EAS}}$ (i.e., $\mathbb{E}[(\hat{I}_{\mathrm{EAS}} - I(\theta))^2]$ is smaller). Thus, EAS yields a statistically more reliable estimate of $I(\theta)$, suggesting a coverage ratio $C_{\mathrm{EAS}}(\theta)$ closer to 1. This fundamental advantage in estimation quality extends to the gradients crucial for optimization.

Next, we analyze the variance of the stochastic gradients used during training. The following theorem shows that EAS leads to a reduced gradient variance compared to GUS.

\begin{theorem}[Gradient Variance Reduction with EAS]
\label{thm:gradient_variance_reduction_main}
Let $G_{\text{EAS}}(\theta) = \nabla_\theta \hat{I}_{\text{EAS}}(\theta)$ and $G_{\text{GUS}}(\theta) = \nabla_\theta \hat{I}_{\text{GUS}}(\theta)$ be the stochastic gradient estimators of $\nabla_\theta I(\theta)$ derived from EAS and GUS based loss formulations, respectively. Under the same assumptions as Theorem~\ref{thm:variance_reduction}, EAS yields a gradient estimator with lower (or equal) total variance: \(\mathrm{Tr}(\mathrm{Cov}(G_{\mathrm{EAS}}(\theta))) \le \mathrm{Tr}(\mathrm{Cov}(G_{\mathrm{GUS}}(\theta)))\).
\end{theorem}
\noindent
\textit{Proof Sketch.} The variance of each gradient component estimator under EAS and GUS is calculated. The comparison hinges on the inequality $\sum_{k=1}^K |E_k| \mu_{g_p,k}^2 \ge |\Omega| \mu_{g_p}^2$ (where $\mu_{g_p,k}$ and $\mu_{g_p}$ are mean gradient components in element $E_k$ and domain $\Omega$, respectively), which follows from Jensen's inequality. This extends to all components, and thus the trace. (Complete proof in Appendix~\ref{app:comparative_analysis}). \qed

\paragraph{Connection to Set-based Processing:}
Our theoretical analysis (Theorems~\ref{thm:variance_reduction} and \ref{thm:gradient_variance_reduction_main}) reveals that Element-Aware Sampling yields lower-variance estimates for both the PDE residual energy and, critically, its gradients compared to Global Uniform Sampling. SetPINNs are architecturally designed to exploit this fundamental statistical advantage. The prospect of reduced gradient variance suggests a more stable optimization signal \citep{bottou2018optimization}, which can theoretically lead to more efficient and reliable convergence. Concurrently, by processing element-wise point sets where this sampling strategy ensures more dependable local physical information (a consequence of Theorem~\ref{thm:variance_reduction}), the SetPINN's attention mechanism is poised to more effectively learn local dependencies. This synergy between superior statistical estimation from sampling and a dependency-aware architecture provides a strong theoretical basis for SetPINNs to accurately resolve local PDE features and achieve robust performance, which we will evaluate empirically in Section~\ref{sec:experiments}.

\begin{figure*}[t]
    \begin{center}
    \includegraphics[scale=0.73]{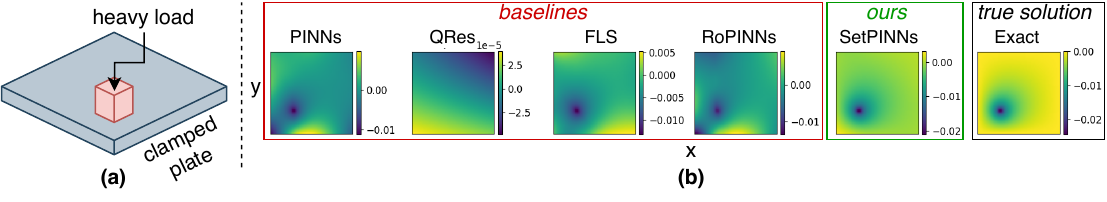}
    \caption{Setup and performance comparison for a clamped plate with a localized heavy load. (a) Illustration of the plate and load placement, representing a common structural engineering scenario. (b) Comparison of SetPINNs and baseline methods against the true solution, highlighting SetPINNs’ ability to capture high-stress regions due to element-aware sampling and a set-based architecture.}
    \label{fig:plate}
    \end{center}
\end{figure*}

\section{Experiments}
\label{sec:experiments}

To empirically demonstrate the effectiveness of SetPINNs, we evaluate its performance against baseline methods on both standard PDE benchmarks (Section~\ref{sec:benchmark}) and real-world tasks (Section~\ref{sec:real-world}).

\paragraph{Baselines.} 
We compare SetPINNs against several strong baselines, including standard MLP-based PINNs \citep{raissi2019physics}, First-Layer Sine networks (FLS) \citep{wong2022learning}, Quadratic Residual Networks (QRes) \citep{bu2021quadratic}, PINNsFormer \citep{zhao2024pinnsformer}, and RoPINNs \citep{wu2024ropinn}, covering recent advances in both architectural design and domain-aware modeling. All models are trained with approximately equal parameter counts to ensure fair comparison. Performance is evaluated using the relative Root Mean Squared Error (rRMSE) \citep{wu2024ropinn}, a standard metric for physics-informed regression.

\paragraph{Implementation Details.}
For baseline models, we use the authors’ official open-source implementations. SetPINNs employ a simple encoder-only Transformer \citep{vaswani2017attention} to capture local set-level dependencies, trained using the objective in Equation~\ref{eq:setpinns_loss_energy} with weighting $\lambda_X = 1$. All models are trained on NVIDIA A100 GPUs. The number of collocation points ($M$) is kept the same across all methods. Training proceeds in two stages: Adam \citep{kingma2014adam} for initial optimization, followed by L-BFGS \citep{schmidt2005minfunc} for fine-tuning. Complete training and hyperparameter details are provided in Appendix~\ref{app:hyperparameters}.

\subsection{White-box PDE Benchmarks}
\label{sec:benchmark}

\paragraph{Setup.} We test SetPINNs on diverse white-box PDE benchmarks: 1D reaction-diffusion, 1D wave, convection, Navier-Stokes, harmonic Poisson, a clamped plate, and the 3D Helmholtz equation. These benchmarks present unique modeling challenges, including sharp gradients (reaction, wave), false diffusion (convection), coupled nonlinearities (Navier-Stokes), high-frequency oscillations (Helmholtz), and boundary sensitivity (harmonic). The domain \(\Omega\) for each task is discretized into a $50 \times 50$ grid, subsequently partitioned into $2 \times 2$ square elements. SetPINNs sample 4 points per element for set-based inputs; baselines use the same total points via global uniform sampling. All models use identical data splits (denser test sets) and are trained purely from physics constraints (PDEs, ICs/BCs). PINNsFormer results are N/A for two equations due to the absence of a temporal dimension. PDE setup and failure modes are detailed in Appendix~\ref{app:pde_setup}.

\paragraph{Qualitative Evaluation.} A particularly challenging case is the clamped plate scenario from structural engineering, where a pinned plate is subjected to a concentrated load at its center \cite{timoshenko1959theory, harris1999structural}. This setup induces highly localized stress, making it a strong test of a model’s ability to resolve sharp spatial features. Figure~\ref{fig:plate} shows that standard PINNs and other baselines struggle in the high-stress region as well as boundary due to sparse or uneven supervision, whereas SetPINNs capture the stress concentration more precisely. This improvement arises from two key aspects: first, element-aware sampling prevents under-sampling in critical zones by ensuring uniform representation across the domain; second, the set-based architecture allows the model to capture local PDE structure more faithfully. This aligns with our theoretical findings that element-aware sampling reduces estimator variance, validating its practical advantage in resolving sharp localized features.

\paragraph{Quantitative Evaluation.} In Table~\ref{tab:overall} we report the relative root mean squared error (rRMSE) averaged over ten independent runs, with associated variance and statistical significance (via two-tailed t-tests). SetPINNs consistently outperform baseline methods across most tasks, achieving lower errors and reduced variance. Improvements are particularly pronounced on tasks that involve localized features, sharp gradients, or under-resolved dynamics, demonstrating the robustness and accuracy of the SetPINN framework in solving challenging physical systems.

\begin{table*}[t]
\caption{SetPINNs as a robust and accurate PDE solver across white-box and grey-box settings. rRMSE scores are reported for each task and model, averaged over ten runs. Statistically significant results (p-value $<$ 0.05) are highlighted in \fcolorbox{calmblue}{calmblue}{blue}, and second-best in \fcolorbox{lighterblue!50}{lighterblue!50}{light blue}.}
\label{tab:overall}
\setlength{\tabcolsep}{0.2pt}
\begin{center}
\begin{small}
\begin{tabular}{
    l | 
    >{\centering\arraybackslash}m{2.0cm} 
    >{\centering\arraybackslash}m{2.0cm} 
    >{\centering\arraybackslash}m{2.0cm} 
    >{\centering\arraybackslash}m{2.0cm} 
    >{\centering\arraybackslash}m{2.0cm} 
    >{\centering\arraybackslash}m{2.0cm}
}
\toprule
\textbf{Task} & 
\makecell{\textbf{SetPINNs}\\\textbf{(ours)}} & 
\makecell{PINNs\\\textcolor{gray}{(JCP'19)}} & 
\makecell{QRes\\\textcolor{gray}{(ICDM'21)}} & 
\makecell{FLS\\\textcolor{gray}{(TAI'22)}} & 
\makecell{PINNsFormer\\\textcolor{gray}{(ICLR'24)}} & 
\makecell{RoPINNs\\\textcolor{gray}{(NeurIPS'24)}} \\
\midrule
\multicolumn{7}{c}{\textbf{White-box PDE Benchmarks}} \\
\midrule
1D-Wave & \cellcolor{calmblue}$0.078 \pm 0.02$ & $0.148 \pm 0.01$ & $0.561 \pm 0.05$ & $0.480 \pm 0.19$ & $0.436 \pm 0.00$ & \cellcolor{lighterblue!50}$0.108 \pm 0.02$ \\
1D-Reaction & \cellcolor{lighterblue!50}$0.061 \pm 0.00$ & $0.801 \pm 0.12$ & $0.989 \pm 0.00$ & $0.883 \pm 0.07$ & $0.505 \pm 0.22$ & \cellcolor{calmblue}$0.030 \pm 0.00$ \\
Convection & \cellcolor{calmblue}$0.031 \pm 0.00$ & $1.136 \pm 0.13$ & $1.004 \pm 0.03$ & $0.897 \pm 0.28$ & \cellcolor{lighterblue!50}$0.650 \pm 0.12$ & $0.961 \pm 0.06$ \\
3D-Helmholtz & \cellcolor{calmblue}$0.072 \pm 0.02$ & $0.281 \pm 0.08$ & $0.217 \pm 0.05$ & \cellcolor{lighterblue!50}$0.189 \pm 0.04$ & $\texttt{N/A}$ & $0.132 \pm 0.03$ \\
Navier-Stokes & \cellcolor{calmblue}$0.218 \pm 0.08$ & $5.703 \pm 2.46$ & $2.235 \pm 1.34$ & $3.016 \pm 0.91$ & \cellcolor{lighterblue!50}$0.841 \pm 0.29$ & $0.935 \pm 0.53$ \\
Harmonic & \cellcolor{calmblue}$0.025 \pm 0.00$ & $0.342 \pm 0.15$ & $0.162 \pm 0.18$ & $0.123 \pm 0.08$ & $0.109 \pm 0.12$ & \cellcolor{lighterblue!50}$0.092 \pm 0.04$ \\
Plate & \cellcolor{calmblue}$0.324 \pm 0.05$ & $1.467 \pm 0.58$ & \cellcolor{lighterblue!50}$1.003 \pm 0.00$ & $1.234 \pm 0.37$ & $\texttt{N/A}$ & $1.375 \pm 0.44$ \\
\midrule
\multicolumn{7}{c}{\textbf{Grey-box Benchmarks}} \\
\midrule
Act. Coeff. & \cellcolor{calmblue}$0.090 \pm 0.00$ & $0.158 \pm 0.00$ & $0.169 \pm 0.00$ & \cellcolor{lighterblue!50}$0.152 \pm 0.00$ & $\texttt{N/A}$ & $\texttt{N/A}$ \\
Agg. Break. & \cellcolor{calmblue}$0.220 \pm 0.01$ & \cellcolor{lighterblue!50}$0.451 \pm 0.00$ & $0.512 \pm 0.03$ & \cellcolor{lighterblue!50}$0.451 \pm 0.00$ & $\texttt{N/A}$ & $\texttt{N/A}$ \\
\bottomrule
\end{tabular}
\end{small}
\end{center}
\end{table*}

\subsection{Grey-box (Real World) Chemical Process Engineering Tasks}
\label{sec:real-world}

\paragraph{Setup.}
We address two chemical process engineering tasks: predicting activity coefficients (AC), constrained by Gibbs-Duhem, and modeling agglomerate breakage (AB), governed by population balances. These tasks use sparse, real-world data (e.g., AC from DDB \citep{DDB2023}; AB via analytical kernels) that irregularly cover the domain. Critically, these are permutation-invariant systems regarding their constituent molecules/particles. Standard PINNs, requiring fixed-size, ordered inputs, struggle with such variable, unordered sets. SetPINNs inherently handle this via permutation-equivariant set processing. Physics-based models like UNIFAC \citep{Fredenslund1975} are consistent but less flexible. Methods such as PINNsFormer and RoPINNs are ill-suited for grey-box scenarios, and hence N/A.

\paragraph{Results.}
Table~\ref{tab:overall} shows SetPINNs achieve the lowest rRMSE and variance on both tasks. For AC, SetPINNs (0.090 ± 0.00) enforce Gibbs-Duhem and outperform UNIFAC (0.166). This highlights SetPINNs' advantage in learning from sparse, irregular real-world data in permutation-invariant chemical systems while respecting physical laws.

\section{Discussion}
\label{sec:discussions}
We analyze the factors behind SetPINNs' improved performance through additional experiments and highlight limitations and future directions.

\begin{figure*}[t]
\begin{center}
    \includegraphics[scale=0.15]{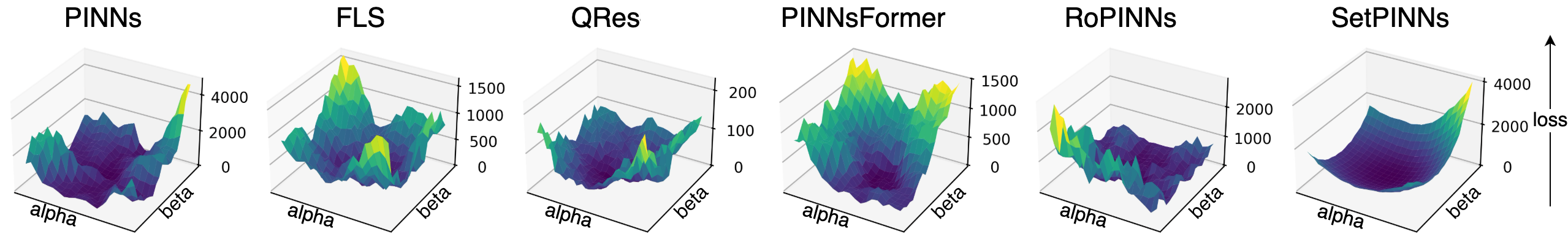}
    \caption{Loss landscape of SetPINNs compared to baselines. Baseline models exhibit sharp cones in their loss landscape, whereas SetPINNs demonstrate a significantly smoother loss surface. This indicates that SetPINNs optimize more stably, reducing sensitivity to failure modes.}
    \label{fig:val_loss}
\end{center}
\end{figure*}

\paragraph{Smoother Loss Landscape and Stable Optimization.}
SetPINNs exhibit smoother loss landscapes (Figure~\ref{fig:val_loss}) when visualized along dominant Hessian directions \cite{li2018visualizing,krishnapriyan2021characterizing}. This characteristic suggests enhanced optimization stability, facilitating better gradient flow and more consistent convergence. Theoretically, this is supported by our variance reduction result (Theorem~\ref{thm:gradient_variance_reduction_main}), which shows that EAS reduces the total variance of gradient estimates. Such smoothness can act as an implicit regularizer, potentially reducing sensitivity to hyperparameter choices.

\begin{figure*}[t]
\begin{center}
    \includegraphics[scale=0.23]{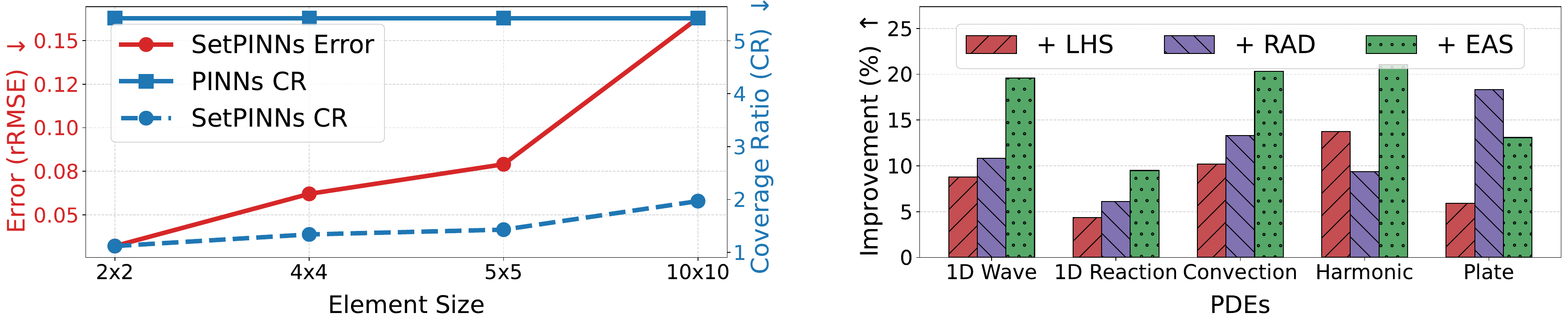}
    \caption{SetPINN design and sampling analysis. Left: SetPINN error and Coverage Ratio (CR) vs. element size. All tested SetPINN configurations outperform the standard PINN baseline in both error and CR. Smaller SetPINN elements further improves error and CR by reducing intra-set distances. Right: Relative rRMSE improvement (\%) of vanilla PINNs with advanced sampling (EAS, LHS, RAD) over uniform sampling. Proposed EAS typically provides the largest benefits.}
    \label{fig:cr_sampling}
\end{center}
\end{figure*}

\paragraph{Impact of Element Size and EAS.}
Domain partitioning is central to the SetPINNs framework. To assess its effect, we evaluate how varying element sizes influence prediction accuracy on the reaction-diffusion equation. The domain is discretized into a $100 \times 100$ grid and partitioned into $n \times n$ square elements, where each element forms a point set via the set generator. We sample \(m_k=n^2\) points from each element to preserve the proportional allocation. As shown in Figure~\ref{fig:cr_sampling} (left), larger elements result in higher errors due to greater inter-point distances, which weaken local interactions. In contrast, smaller elements better preserve locality, leading to improved accuracy.

We also study the effectiveness of Element-Aware Sampling (EAS) in Figure~\ref{fig:cr_sampling} (right), comparing relative rRMSE improvements of vanilla PINNs with various sampling strategies: Latin Hypercube Sampling (LHS), Residual-based Adaptive Distribution (RAD) \cite{wu2023comprehensive}, and EAS, over uniform sampling. LHS is a widely used stratified method, while RAD adaptively focuses on high-residual regions. Among these, EAS yields the greatest performance gain, highlighting its contribution to improved supervision. Further accuracy gains beyond sampling are attributed to the set-based architecture. Additional ablation results are provided in Appendix~\ref{app:ablation}.

\paragraph{Limitations and Future Directions.}
This paper presents SetPINNs as a novel framework for physics-informed learning, with both theoretical grounding and strong empirical results. However, several aspects merit further investigation. Our theoretical analysis of EAS assumes uniform sampling within elements and proportional allocation. While our experiments already include irregular domains, exploring non-uniform or adaptive sampling strategies could provide deeper insights. SetPINNs also introduce hyperparameters tied to domain partitioning (e.g., element number and size) and the set architecture (e.g., Transformer depth, attention heads). Although our ablations (Appendix~\ref{app:ablation}) and runtime analysis (Appendix~\ref{app:compute}) demonstrate robust trends, optimal configurations may depend on the PDE. Finally, while our complexity analysis (Appendix~\ref{app:computation_complexity}) shows SetPINNs scale independently of domain dimensionality, and we validate this on high-dimensional problems (e.g., 3D Helmholtz), future work could explore more efficient set processing for very high-dimensional settings.

\section{Conclusion}
\label{sec:conclusion}
By integrating element-aware sampling and set-based learning, SetPINNs offer a principled way to capture local PDE features. Theoretical analysis demonstrates that this combination reduces both estimator and gradient variance, leading to improved residual coverage and more stable optimization. Across a wide range of experiments, the method demonstrates superior accuracy and stability over conventional PINNs and state-of-the-art variants. In doing so, it presents a compelling direction for the next generation of physics-informed neural networks, suggesting that explicitly modeling local dependencies in the domain can yield substantial gains in both performance and reliability.

\bibliographystyle{plainnat}
\bibliography{paper}

\newpage
\appendix
\section{Detailed Theoretical Analysis}
\label{app:theoretical_analysis}

This appendix provides detailed proofs for the theoretical results presented in Section~\ref{sec:theory}, establishing the statistical advantages of Element-Aware Sampling (EAS) over Global Uniform Sampling (GUS) for estimating the total squared PDE residual integral.

\subsection{Assumptions}
Our theoretical analysis relies on the following assumptions regarding the domain partitioning, sampling strategies, and the PDE residual:
\begin{enumerate}
    \item \textbf{Domain Partition:} The domain $\Omega \subset \mathbb{R}^d$ is partitioned into $K$ non-overlapping measurable elements $\{E_k\}_{k=1}^K$ such that $\bigcup_{k=1}^K E_k = \Omega$ and $|E_i \cap E_j| = 0$ for $i \neq j$.
    \item \textbf{Uniform Random Sampling (EAS):} For Element-Aware Sampling, $m_k$ points $\{x_{k,i}\}_{i=1}^{m_k}$ are independently and uniformly sampled from each element $E_k$.
    \item \textbf{Uniform Random Sampling (GUS):} For Global Uniform Sampling, $M$ points $\{x_j\}_{j=1}^M$ are independently and uniformly sampled from the entire domain $\Omega$.
    \item \textbf{Proportional Sampling Allocation:} The number of samples per element in EAS is proportional to the element's measure, i.e., $m_k = M \frac{|E_k|}{|\Omega|}$ for all $k=1,\dots,K$, where $M = \sum_{k=1}^K m_k$ is the total number of samples.
    \item \textbf{Residual Integrability:} The squared residual function $|R_\theta(x)|^2$ is integrable over $\Omega$, i.e., $\int_\Omega |R_\theta(x)|^2 dx < \infty$. For variance calculations, we further assume $|R_\theta(x)|^4$ is integrable over $\Omega$, i.e., $\int_\Omega |R_\theta(x)|^4 dx < \infty$.
\end{enumerate}

Let $\phi(x) = |R_\theta(x)|^2$ for brevity in the proofs. The total integral is $I = \int_\Omega \phi(x) dx$. The estimators for this integral are:
\begin{align*}
    \hat{I}_{\mathrm{GUS}} &= \frac{|\Omega|}{M} \sum_{j=1}^M \phi(x_j) \\
    \hat{I}_{\mathrm{EAS}} &= \sum_{k=1}^K \frac{|E_k|}{m_k} \sum_{i=1}^{m_k} \phi(x_{k,i})
\end{align*}
Note that under Assumption 4 (Proportional Allocation), the EAS estimator can also be written as:
\[
\hat{I}_{\mathrm{EAS}} = \sum_{k=1}^K \frac{|E_k|}{M |E_k| / |\Omega|} \sum_{i=1}^{m_k} \phi(x_{k,i}) = \frac{|\Omega|}{M} \sum_{k=1}^K \sum_{i=1}^{m_k} \phi(x_{k,i}).
\]
We will use this second form for EAS when relying on proportional allocation, as specified in the Lemma and Theorem statements.

\subsection{Proofs and Statistical Analysis on Coverage}
\label{app:proof_coverage}

\begin{proof}[\textbf{Proof of Lemma~\ref{lem:unbiased_estimation}}]
We aim to show that $\mathbb{E}[\hat{I}_{\mathrm{GUS}}] = I$ and $\mathbb{E}[\hat{I}_{\mathrm{EAS}}] = I$.

\paragraph{Global Uniform Sampling (GUS):}
The points $x_j$ are sampled independently and uniformly from $\Omega$ (Assumption 3). The expected value of $\phi(x_j)$ for any single point $x_j \sim U(\Omega)$ is:
\[
\mathbb{E}[\phi(x_j)] = \int_\Omega \phi(x) p(x) dx = \int_\Omega \phi(x) \frac{1}{|\Omega|} dx = \frac{1}{|\Omega|} I.
\]
Using the linearity of expectation for the estimator $\hat{I}_{\mathrm{GUS}}$:
\begin{align*}
\mathbb{E}[\hat{I}_{\mathrm{GUS}}] &= \mathbb{E}\left[ \frac{|\Omega|}{M} \sum_{j=1}^{M} \phi(x_j) \right] \\
&= \frac{|\Omega|}{M} \sum_{j=1}^{M} \mathbb{E}[\phi(x_j)] \\
&= \frac{|\Omega|}{M} \sum_{j=1}^{M} \left( \frac{1}{|\Omega|} I \right) \\
&= \frac{|\Omega|}{M} \cdot M \cdot \frac{I}{|\Omega|} = I.
\end{align*}
Thus, $\hat{I}_{\mathrm{GUS}}$ is an unbiased estimator of $I$.

\paragraph{Element-Aware Sampling (EAS):}
We use the general definition $\hat{I}_{\mathrm{EAS}} = \sum_{k=1}^K \frac{|E_k|}{m_k} \sum_{i=1}^{m_k} \phi(x_{k,i})$. The points $x_{k,i}$ are sampled independently and uniformly from the element $E_k$ (Assumption 2). The expected value of $\phi(x_{k,i})$ for a point within element $E_k$ is:
\[
\mathbb{E}[\phi(x_{k,i})] = \int_{E_k} \phi(x) p(x | x \in E_k) dx = \int_{E_k} \phi(x) \frac{1}{|E_k|} dx = \frac{1}{|E_k|} \int_{E_k} \phi(x) dx.
\]
Let $I_k = \int_{E_k} \phi(x) dx$. Then $\mathbb{E}[\phi(x_{k,i})] = I_k / |E_k|$.
Using the linearity of expectation for $\hat{I}_{\mathrm{EAS}}$:
\begin{align*}
\mathbb{E}[\hat{I}_{\mathrm{EAS}}] &= \mathbb{E}\left[ \sum_{k=1}^K \frac{|E_k|}{m_k} \sum_{i=1}^{m_k} \phi(x_{k,i}) \right] \\
&= \sum_{k=1}^K \frac{|E_k|}{m_k} \sum_{i=1}^{m_k} \mathbb{E}[\phi(x_{k,i})] \\
&= \sum_{k=1}^K \frac{|E_k|}{m_k} \sum_{i=1}^{m_k} \left( \frac{I_k}{|E_k|} \right) \\
&= \sum_{k=1}^K \frac{|E_k|}{m_k} m_k \left( \frac{I_k}{|E_k|} \right) \\
&= \sum_{k=1}^K I_k.
\end{align*}
Since the elements $\{E_k\}_{k=1}^K$ form a partition of $\Omega$ (Assumption 1), the sum of integrals over the elements equals the integral over the entire domain:
\[
\mathbb{E}[\hat{I}_{\mathrm{EAS}}] = \sum_{k=1}^K \int_{E_k} \phi(x) dx = \int_\Omega \phi(x) dx = I.
\]
Thus, $\hat{I}_{\mathrm{EAS}}$ is also an unbiased estimator of $I$, regardless of the allocation $m_k$ (as long as $m_k > 0$). The Lemma statement in the main text implicitly assumed proportional allocation for simplicity, but the result holds more generally. For consistency with the Theorem, we rely on proportional allocation moving forward.
\end{proof}

\begin{proof}[\textbf{Proof of Theorem~\ref{thm:variance_reduction}}]
We compute the variances for both estimators, assuming proportional allocation for EAS (Assumption 4), i.e., $m_k = M |E_k| / |\Omega|$, and using the corresponding simplified form $\hat{I}_{\mathrm{EAS}} = \frac{|\Omega|}{M} \sum_{k=1}^K \sum_{i=1}^{m_k} \phi(x_{k,i})$. Let $\mu_k = \mathbb{E}[\phi(x)]$ for $x \sim U(E_k)$, so $\mu_k = I_k / |E_k|$.

\paragraph{Variance for Global Uniform Sampling (GUS):}
Since the samples $x_j$ are i.i.d. $U(\Omega)$ (Assumption 3):
\begin{align*}
\mathrm{Var}(\hat{I}_{\mathrm{GUS}}) &= \mathrm{Var}\left( \frac{|\Omega|}{M} \sum_{j=1}^{M} \phi(x_j) \right)
= \left(\frac{|\Omega|}{M}\right)^2 \sum_{j=1}^{M} \mathrm{Var}(\phi(x_j)) \\
&= \frac{|\Omega|^2}{M} \mathrm{Var}(\phi(x_1)) \\
&= \frac{|\Omega|^2}{M} \left( \mathbb{E}[\phi(x_1)^2] - (\mathbb{E}[\phi(x_1)])^2 \right) \\
&= \frac{|\Omega|^2}{M} \left( \frac{1}{|\Omega|} \int_\Omega \phi(x)^2 dx - \left( \frac{I}{|\Omega|} \right)^2 \right) \\
&= \frac{|\Omega|}{M} \int_\Omega \phi(x)^2 dx - \frac{1}{M} I^2. \label{eq:app_var_gus} \tag{A.1}
\end{align*}
This requires Assumption 5 ($\int \phi^2 dx < \infty$).

\paragraph{Variance for Element-Aware Sampling (EAS) with Proportional Allocation:}
Let $Y_k = \sum_{i=1}^{m_k} \phi(x_{k,i})$. Then $\hat{I}_{\mathrm{EAS}} = \frac{|\Omega|}{M} \sum_{k=1}^K Y_k$. Since samples from different elements are independent, the $Y_k$ are independent random variables.
\begin{align*}
\mathrm{Var}(\hat{I}_{\mathrm{EAS}}) &= \mathrm{Var}\left( \frac{|\Omega|}{M} \sum_{k=1}^K Y_k \right)
= \left(\frac{|\Omega|}{M}\right)^2 \sum_{k=1}^K \mathrm{Var}(Y_k).
\end{align*}
Within element $E_k$, the samples $x_{k,i}$ are i.i.d. $U(E_k)$ (Assumption 2).
\begin{align*}
\mathrm{Var}(Y_k) &= \mathrm{Var}\left( \sum_{i=1}^{m_k} \phi(x_{k,i}) \right) = \sum_{i=1}^{m_k} \mathrm{Var}(\phi(x_{k,i})) = m_k \mathrm{Var}(\phi(x_{k,1})).
\end{align*}
The variance of $\phi(x_{k,1})$ for $x_{k,1} \sim U(E_k)$ is:
\begin{align*}
\mathrm{Var}(\phi(x_{k,1})) &= \mathbb{E}[\phi(x_{k,1})^2] - (\mathbb{E}[\phi(x_{k,1})])^2 \\
&= \left( \frac{1}{|E_k|} \int_{E_k} \phi(x)^2 dx \right) - \mu_k^2.
\end{align*}
So, $\mathrm{Var}(Y_k) = m_k \left( \frac{1}{|E_k|} \int_{E_k} \phi(x)^2 dx - \mu_k^2 \right)$.
Substituting this and the proportional allocation $m_k = M \frac{|E_k|}{|\Omega|}$ (Assumption 4):
\begin{align*}
\mathrm{Var}(\hat{I}_{\mathrm{EAS}}) &= \left(\frac{|\Omega|}{M}\right)^2 \sum_{k=1}^K m_k \left( \frac{1}{|E_k|} \int_{E_k} \phi(x)^2 dx - \mu_k^2 \right) \\
&= \left(\frac{|\Omega|}{M}\right)^2 \sum_{k=1}^K \left( M \frac{|E_k|}{|\Omega|} \right) \left( \frac{1}{|E_k|} \int_{E_k} \phi(x)^2 dx - \mu_k^2 \right) \\
&= \frac{|\Omega|}{M} \sum_{k=1}^K |E_k| \left( \frac{1}{|E_k|} \int_{E_k} \phi(x)^2 dx - \mu_k^2 \right) \\
&= \frac{|\Omega|}{M} \sum_{k=1}^K \left( \int_{E_k} \phi(x)^2 dx - |E_k| \mu_k^2 \right) \\
&= \frac{|\Omega|}{M} \left( \sum_{k=1}^K \int_{E_k} \phi(x)^2 dx \right) - \frac{|\Omega|}{M} \sum_{k=1}^K |E_k| \mu_k^2 \\
&= \frac{|\Omega|}{M} \int_\Omega \phi(x)^2 dx - \frac{|\Omega|}{M} \sum_{k=1}^K |E_k| \left( \frac{I_k}{|E_k|} \right)^2 \\
&= \frac{|\Omega|}{M} \int_\Omega \phi(x)^2 dx - \frac{|\Omega|}{M} \sum_{k=1}^K \frac{I_k^2}{|E_k|}. \label{eq:app_var_eas} \tag{A.2}
\end{align*}
This also requires Assumption 5.

\paragraph{Comparing Variances:}
We want to show that $\mathrm{Var}(\hat{I}_{\mathrm{EAS}}) \le \mathrm{Var}(\hat{I}_{\mathrm{GUS}})$. Using equations \eqref{eq:app_var_eas} and \eqref{eq:app_var_gus}, this is equivalent to showing:
\[
\frac{|\Omega|}{M} \int_\Omega \phi(x)^2 dx - \frac{|\Omega|}{M} \sum_{k=1}^K \frac{I_k^2}{|E_k|} \le \frac{|\Omega|}{M} \int_\Omega \phi(x)^2 dx - \frac{1}{M} I^2.
\]
This simplifies to:
\[
- \frac{|\Omega|}{M} \sum_{k=1}^K \frac{I_k^2}{|E_k|} \le - \frac{1}{M} I^2.
\]
Multiplying by $-M/|\Omega|$ (and reversing the inequality sign):
\[
\sum_{k=1}^K \frac{I_k^2}{|E_k|} \ge \frac{I^2}{|\Omega|}.
\]
Recall $I = \sum_{k=1}^K I_k$ and $|\Omega| = \sum_{k=1}^K |E_k|$. Let $w_k = |E_k|$. The inequality becomes:
\[
\sum_{k=1}^K \frac{I_k^2}{w_k} \ge \frac{(\sum_{k=1}^K I_k)^2}{\sum_{k=1}^K w_k}.
\]
This inequality follows directly from the Cauchy-Schwarz inequality. Let vectors $a = (\sqrt{w_1}, \dots, \sqrt{w_K})$ and $b = (I_1/\sqrt{w_1}, \dots, I_K/\sqrt{w_K})$. Then $(\sum a_i b_i)^2 \le (\sum a_i^2) (\sum b_i^2)$, which translates to:
\[
\left( \sum_{k=1}^K \sqrt{w_k} \cdot \frac{I_k}{\sqrt{w_k}} \right)^2 \le \left( \sum_{k=1}^K w_k \right) \left( \sum_{k=1}^K \frac{I_k^2}{w_k} \right)
\]
\[
\left( \sum_{k=1}^K I_k \right)^2 \le \left( \sum_{k=1}^K w_k \right) \left( \sum_{k=1}^K \frac{I_k^2}{w_k} \right)
\]
Rearranging gives the desired inequality:
\[
\sum_{k=1}^K \frac{I_k^2}{w_k} \ge \frac{(\sum I_k)^2}{\sum w_k}.
\]
Therefore, $\mathrm{Var}(\hat{I}_{\mathrm{EAS}}) \le \mathrm{Var}(\hat{I}_{\mathrm{GUS}})$. Equality holds if and only if $b$ is proportional to $a$, which means $I_k / \sqrt{w_k} = c \sqrt{w_k}$ for some constant $c$, implying $I_k / w_k = c$. This means the average value of $\phi(x)$ is the same in all elements $E_k$.
\end{proof}

\begin{corollary}[Improved Reliability of Coverage Metric]
\label{cor:reliability}
The Element-Aware Sampling strategy provides a more reliable Monte Carlo estimate of the total squared PDE residual integral compared to global uniform sampling. This statistical reliability translates to the coverage ratio $C_{\mathrm{EAS}}(\theta)$ being statistically more likely to be closer to 1 than $C_{\mathrm{GUS}}(\theta)$.
\end{corollary}

\begin{proof}[\textbf{Proof of Corollary~\ref{cor:reliability}}]
Lemma~\ref{lem:unbiased_estimation} established that both estimators $\hat{I}_{\mathrm{EAS}}$ and $\hat{I}_{\mathrm{GUS}}$ are unbiased for the true integral $I = \int_\Omega \phi(x) dx$, under the respective sampling assumptions (and proportional allocation for EAS as used in the theorem).
\[
\mathbb{E}[\hat{I}_{\mathrm{EAS}}] = \mathbb{E}[\hat{I}_{\mathrm{GUS}}] = I.
\]
Theorem~\ref{thm:variance_reduction} established that under proportional allocation, the variance of the EAS estimator is less than or equal to the variance of the GUS estimator:
\[
\mathrm{Var}(\hat{I}_{\mathrm{EAS}}) \le \mathrm{Var}(\hat{I}_{\mathrm{GUS}}).
\]
An estimator's reliability is related to its precision, often measured by its variance. An unbiased estimator with lower variance is generally considered more reliable because its distribution is more concentrated around the true value it estimates. This means that for any given realization of the sampling process, the estimate $\hat{I}_{\mathrm{EAS}}$ is statistically more likely to be close to the true value $I$ than the estimate $\hat{I}_{\mathrm{GUS}}$.

The coverage ratios (Definition~\ref{def:coverage_ratio}) are $C_{\mathrm{GUS}} = I / \hat{I}_{\mathrm{GUS}}$ and $C_{\mathrm{EAS}} = I / \hat{I}_{\mathrm{EAS}}$. Since $\hat{I}_{\mathrm{EAS}}$ is a statistically more reliable estimator of $I$ (i.e., has lower variance around $I$) compared to $\hat{I}_{\mathrm{GUS}}$, the resulting ratio $C_{\mathrm{EAS}}$ is statistically more likely to be close to $I/I = 1$ than $C_{\mathrm{GUS}}$. Thus, EAS provides a more reliable assessment of the residual coverage compared to GUS.
\end{proof}

\subsection{Comparative Statistical Analysis of SetPINN (EAS) and Standard PINN (GUS) Loss Formulations}
\label{app:comparative_analysis}

This section provides a direct statistical comparison between the loss function formulation underpinning SetPINNs, which leverages Element-Aware Sampling (EAS), and the loss function typical of standard PINNs, which uses Global Uniform Sampling (GUS). We analyze the properties of these loss estimators and their gradients.

Let $u_\theta(x)$ be the neural network approximation of the PDE solution, parameterized by $\theta$. Let $\phi(x; \theta) = \| \mathcal{O}_\Omega(u_\theta(x)) \|^2_2$ be the squared PDE residual at point $x \in \Omega$ (for simplicity, considering the primary PDE residual operator; the argument extends to other constraint terms). The total true residual energy over the domain $\Omega$ is $I(\theta) = \int_{\Omega} \phi(x; \theta) dx$.

\subsubsection{Loss Function Estimators}

We consider the loss functions as estimators of the total residual energy $I(\theta)$, consistent with Definition~\ref{def:coverage_ratio} in the main paper.

\paragraph{1. Standard PINN with Global Uniform Sampling (GUS)}
The loss is an estimator of $I(\theta)$ based on $M$ points $\{x_j\}_{j=1}^M$ sampled i.i.d. from $U(\Omega)$:
\begin{equation}
    L_{\text{GUS}}(\theta; \{x_j\}) \equiv \hat{I}_{\text{GUS}}(\theta) = \frac{|\Omega|}{M} \sum_{j=1}^M \phi(x_j; \theta).
    \label{eq:app_b_L_gus}
\end{equation}
As per Lemma~\ref{lem:unbiased_estimation}, $\mathbb{E}[\hat{I}_{\text{GUS}}(\theta)] = I(\theta)$.
The variance is derived in Appendix~\ref{app:theoretical_analysis} (Equation~\eqref{eq:app_var_gus}):
\begin{equation}
    \mathrm{Var}(\hat{I}_{\text{GUS}}(\theta)) = \frac{|\Omega|}{M} \int_\Omega \phi(x; \theta)^2 dx - \frac{1}{M} I(\theta)^2.
    \label{eq:app_b_Var_L_gus}
\end{equation}

\paragraph{2. SetPINN with Element-Aware Sampling (EAS)}
The domain $\Omega$ is partitioned into $K$ disjoint elements $E_k$. A total of $M$ points are sampled, with $m_k$ points $\{x_{k,i}\}_{i=1}^{m_k}$ sampled i.i.d. from $U(E_k)$ for each element $E_k$. Proportional allocation (Assumption 4, Appendix~\ref{app:theoretical_analysis}) states $m_k = M \frac{|E_k|}{|\Omega|}$.
The EAS-based estimator of $I(\theta)$ is (Definition~\ref{def:coverage_ratio}, main paper):
\begin{equation}
    L_{\text{SetPINN}}(\theta; \{\mathbf{x}\}) \equiv \hat{I}_{\text{EAS}}(\theta) = \sum_{k=1}^K \frac{|E_k|}{m_k} \sum_{i=1}^{m_k} \phi(x_{k,i}; \theta).
    \label{eq:app_b_L_eas}
\end{equation}
As per Lemma~\ref{lem:unbiased_estimation}, $\mathbb{E}[\hat{I}_{\text{EAS}}(\theta)] = I(\theta)$.
The variance is derived in Appendix~\ref{app:theoretical_analysis} (Equation~\eqref{eq:app_var_eas}):
\begin{equation}
    \mathrm{Var}(\hat{I}_{\text{EAS}}(\theta)) = \frac{|\Omega|}{M} \int_\Omega \phi(x; \theta)^2 dx - \frac{|\Omega|}{M} \sum_{k=1}^K \frac{I_k(\theta)^2}{|E_k|}.
    \label{eq:app_b_Var_L_eas}
\end{equation}
where $I_k(\theta) = \int_{E_k} \phi(x; \theta) dx$.
Theorem~\ref{thm:variance_reduction} (main paper) establishes that $\mathrm{Var}(\hat{I}_{\text{EAS}}(\theta)) \le \mathrm{Var}(\hat{I}_{\text{GUS}}(\theta))$.

\subsubsection{Stochastic Gradient Estimators}
Let $G_{\text{GUS}}(\theta) = \nabla_\theta L_{\text{GUS}}(\theta; \{x_j\})$ and $G_{\text{SetPINN}}(\theta) = \nabla_\theta L_{\text{SetPINN}}(\theta; \{\mathbf{x}\})$.
Both are unbiased estimators of the true gradient $G_{\text{true}}(\theta) = \nabla_\theta I(\theta)$, assuming interchangeability of expectation and differentiation.
We compare their variances. Let $g(x; \theta) = \nabla_\theta \phi(x; \theta)$.

\paragraph{1. Gradient Variance for GUS}
\begin{equation}
    G_{\text{GUS}}(\theta) = \frac{|\Omega|}{M} \sum_{j=1}^M g(x_j; \theta).
\end{equation}
The covariance matrix of this estimator is:
\begin{equation}
    \mathrm{Cov}(G_{\text{GUS}}(\theta)) = \frac{|\Omega|^2}{M^2} \sum_{j=1}^M \mathrm{Cov}_{x_j \sim U(\Omega)}(g(x_j; \theta)) = \frac{|\Omega|^2}{M} \mathrm{Cov}_{x \sim U(\Omega)}(g(x; \theta)).
    \label{eq:app_b_Cov_G_gus}
\end{equation}
The variance (sum of diagonal elements of the covariance matrix, i.e., trace) is $\mathrm{Tr}(\mathrm{Cov}(G_{\text{GUS}}(\theta)))$.

\paragraph{2. Gradient Variance for SetPINN (EAS)}
\begin{equation}
    G_{\text{SetPINN}}(\theta) = \sum_{k=1}^K \frac{|E_k|}{m_k} \sum_{i=1}^{m_k} g(x_{k,i}; \theta).
\end{equation}
Due to independence of sampling across elements and i.i.d. sampling within elements:
\begin{align}
    \mathrm{Cov}(G_{\text{SetPINN}}(\theta)) &= \sum_{k=1}^K \mathrm{Cov}\left( \frac{|E_k|}{m_k} \sum_{i=1}^{m_k} g(x_{k,i}; \theta) \right) \nonumber \\
    &= \sum_{k=1}^K \left(\frac{|E_k|}{m_k}\right)^2 \sum_{i=1}^{m_k} \mathrm{Cov}_{x_{k,i} \sim U(E_k)}(g(x_{k,i}; \theta)) \nonumber \\
    &= \sum_{k=1}^K \frac{|E_k|^2}{m_k} \mathrm{Cov}_{x \sim U(E_k)}(g(x; \theta)).
    \label{eq:app_b_Cov_G_eas}
\end{align}
The variance is $\mathrm{Tr}(\mathrm{Cov}(G_{\text{SetPINN}}(\theta)))$.

\begin{proof}[\textbf{Proof of Theorem~\ref{thm:gradient_variance_reduction_main}}]
The proof structure is analogous to that of Theorem~\ref{thm:variance_reduction} (variance reduction for the function $\phi(x;\theta)$), applied to each component of the gradient vector $g(x;\theta)$.
Consider a single component $g_p(x;\theta)$ of the gradient vector $g(x;\theta)$ (i.e., derivative with respect to $\theta_p$). We want to show $\mathrm{Var}(G_{\text{SetPINN},p}(\theta)) \le \mathrm{Var}(G_{\text{GUS},p}(\theta))$.
From Equation~\eqref{eq:app_b_Cov_G_gus}, $\mathrm{Var}(G_{\text{GUS},p}(\theta)) = \frac{|\Omega|^2}{M} \mathrm{Var}_{x \sim U(\Omega)}(g_p(x; \theta))$.
From Equation~\eqref{eq:app_b_Cov_G_eas}, $\mathrm{Var}(G_{\text{SetPINN},p}(\theta)) = \sum_{k=1}^K \frac{|E_k|^2}{m_k} \mathrm{Var}_{x \sim U(E_k)}(g_p(x; \theta))$.
Substituting $m_k = M |E_k|/|\Omega|$ into the expression for $\mathrm{Var}(G_{\text{SetPINN},p}(\theta))$:
\[
\mathrm{Var}(G_{\text{SetPINN},p}(\theta)) = \sum_{k=1}^K \frac{|E_k|^2}{M |E_k|/|\Omega|} \mathrm{Var}_{x \sim U(E_k)}(g_p(x; \theta)) = \frac{|\Omega|}{M} \sum_{k=1}^K |E_k| \mathrm{Var}_{x \sim U(E_k)}(g_p(x; \theta)).
\]
We need to show:
\[
\frac{|\Omega|}{M} \sum_{k=1}^K |E_k| \mathrm{Var}_{x \sim U(E_k)}(g_p(x; \theta)) \le \frac{|\Omega|^2}{M} \mathrm{Var}_{x \sim U(\Omega)}(g_p(x; \theta)).
\]
This simplifies to:
\[
\sum_{k=1}^K |E_k| \mathrm{Var}_{x \sim U(E_k)}(g_p(x; \theta)) \le |\Omega| \mathrm{Var}_{x \sim U(\Omega)}(g_p(x; \theta)).
\]
Let $\mu_{g_p,k} = \frac{1}{|E_k|} \int_{E_k} g_p(x; \theta) dx$ and $\mu_{g_p} = \frac{1}{|\Omega|} \int_{\Omega} g_p(x; \theta) dx$.
The inequality is equivalent to (following the steps in the proof of Theorem~\ref{thm:variance_reduction} in Appendix~\ref{app:theoretical_analysis}, but with $g_p$ instead of $\phi$):
\[
\sum_{k=1}^K |E_k| \mu_{g_p,k}^2 \ge |\Omega| \mu_{g_p}^2.
\]
This inequality holds due to Jensen's inequality for the convex function $f(z)=z^2$, as shown in the proof of Theorem~\ref{thm:variance_reduction}.
Since this holds for each component $g_p(x;\theta)$ of the gradient vector $g(x;\theta)$, it holds for their sum of variances, which is the trace of the covariance matrix.
Thus, $\mathrm{Tr}(\mathrm{Cov}(G_{\text{SetPINN}}(\theta))) \le \mathrm{Tr}(\mathrm{Cov}(G_{\text{GUS}}(\theta)))$.
\end{proof}

\subsubsection{Implications for Optimization Dynamics and Solution Quality}

Theorem~\ref{thm:gradient_variance_reduction_main} establishes that SetPINNs' Element-Aware Sampling (EAS) provides lower-variance stochastic gradient estimates than Global Uniform Sampling (GUS) for any given neural network state $\theta$. While SetPINNs optimize their \emph{own} loss function, this inherent statistical advantage of EAS directly impacts their training, leading to:

\begin{itemize}
    \item \textbf{More Stable Optimization:} Lower gradient variance implies more reliable updates~\citep{bottou2018optimization}, fostering stabler training with reduced oscillations and improved convergence. This aligns with empirically smoother loss landscapes (Figure~\ref{fig:val_loss}).

    \item \textbf{Increased Solution Accuracy:} More stable optimization enables precise convergence. Concurrently, EAS yields more reliable local physical information (per Theorem~\ref{thm:variance_reduction}), allowing SetPINN's architecture to learn local dependencies effectively, thereby enhancing overall solution accuracy.

    \item \textbf{Enhanced Robustness:} Reduced gradient variance leads to lower performance variability across runs (Table~\ref{tab:overall}). EAS's balanced sampling mitigates undersampling failures in critical regions and improves the handling of complex PDE features.
\end{itemize}

In essence, Theorem~\ref{thm:gradient_variance_reduction_main} provides a theoretical basis for SetPINNs' superior optimization stability, solution accuracy, and robustness, directly explaining the empirical improvements reported in Section~\ref{sec:experiments}.

The theoretical underpinnings of SetPINNs are established with careful rigor in this section, providing a complete derivation of the statistical advantages that drive its performance.

\section{Training and Evaluation}
\label{app:training_eval}

\subsection{Physics-Informed Neural Networks (PINNs)}
\label{app:pinn-loss}

Physics-Informed Neural Networks (PINNs) approximate the solution of partial differential equations (PDEs) by embedding the physical constraints directly into the loss function. Given a spatio-temporal PDE defined over a domain \( \Omega \subset \mathbb{R}^d \), where \( x = (x_1, \dots, x_{d-1}, t) \) represents the input coordinates, the objective is to find \( u(x) \) satisfying

\begin{equation}
\label{eq:pde_appendix}
\begin{aligned}
    &\mathcal{O}_\Omega(u)(x) = 0, \quad x \in \Omega, \\
    &\mathcal{O}_{\Omega_0}(u)(x) = g(x), \quad x \in \Omega_0, \\
    &\mathcal{O}_{\partial\Omega}(u)(x) = h(x), \quad x \in \partial \Omega.
\end{aligned}
\end{equation}

PINNs approximate \( u(x) \) using a neural network \( u_\theta(x) \), where \( \theta \) represents the learnable parameters. The loss function is designed to enforce the PDE residual, initial conditions (ICs), boundary conditions (BCs), and, optionally, additional data constraints:

\begin{equation}
\begin{aligned}
\mathcal{L}_{\text{PINN}}(u_\theta) =
& \frac{\lambda_{\Omega}}{N_{\Omega}} \sum_{i=1}^{N_{\Omega}} \left\| \mathcal{O}_{\Omega}(u_\theta)(x_{\Omega}^{(i)}) \right\|^2 \\
& + \frac{\lambda_{\Omega_0}}{N_{\Omega_0}} \sum_{i=1}^{N_{\Omega_0}} \left\| u_\theta(x_{\Omega_0}^{(i)}) - g(x_{\Omega_0}^{(i)}) \right\|^2 \\
& + \frac{\lambda_{\partial\Omega}}{N_{\partial\Omega}} \sum_{i=1}^{N_{\partial\Omega}} \left\| u_\theta(x_{\partial\Omega}^{(i)}) - h(x_{\partial\Omega}^{(i)}) \right\|^2 \\
& + \frac{\lambda_{\text{data}}}{N_{\text{data}}} \sum_{i=1}^{N_{\text{data}}} \left\| u_\theta(x_{\text{data}}^{(i)}) - u_{\text{true}}(x_{\text{data}}^{(i)}) \right\|^2.
\end{aligned}
\end{equation}

Here, \( N_X \) denotes the number of collocation points sampled from each region \( X \), and \( \lambda_X \) are weight factors balancing the contributions of different terms. The first term ensures that the neural network satisfies the PDE within \( \Omega \), the second term enforces initial conditions at \( \Omega_0 \), the third term applies the boundary conditions along \( \partial \Omega \), and the last term (if available) incorporates supervised data for improved accuracy. This loss formulation ensures that PINNs provide physically consistent solutions while leveraging data-driven constraints when applicable.

\subsection{Training Algorithm of SetPINNs.}
\label{app:training_setpinns}
The training algorithm for SetPINNs, outlined in Algorithm \ref{alg:setpinns-training}, involves initializing the model parameters for the Set Generator, Mixer Network, Set Processor, and PDE Probe. During each training iteration, the algorithm processes each element \( E_k \) in the discretized domain by sampling points to generate a set \( \mathcal{S}_k \), transforming this set into a high-dimensional representation using the Mixer Network \( p_\psi \), and then processing it with the Set Processor \( f_\phi \). The transformed set representation \( \mathcal{O}_k \) is used by the PDE Probe \( q_\eta \) to predict solutions. The set-wise physics loss, defined in Equation \ref{eq:setpinns_loss_energy}, is computed to ensure adherence to physical laws through the localized residual energy formulation. The model parameters are updated iteratively using the Adam optimizer, followed by fine-tuning with the L-BFGS optimizer, to achieve efficient and accurate convergence. This approach leverages set-based processing and attention mechanisms to simultaneously approximate solutions for multiple points in the domain, improving the robustness and accuracy of the learned physics-informed solutions.

\begin{algorithm}[h]
\caption{Training Algorithm for SetPINNs}
\label{alg:setpinns-training}
\begin{algorithmic}[1]
\REQUIRE Domain \( \Omega \) partitioned into \( K \) elements \( \{E_k\}_{k=1}^K \)
\REQUIRE Set Generator
\REQUIRE Mixer Network \( p_\psi \)
\REQUIRE Set Processor \( f_\phi \)
\REQUIRE PDE Probe \( q_\eta \)
\REQUIRE Hyperparameters \( \lambda_{\Omega} \), \( \lambda_{\Omega_0} \), \( \lambda_{\partial\Omega} \)
\STATE Initialize model parameters \( \psi \), \( \phi \), \( \eta \)
\FOR{each training iteration}
    \FOR{each element \( E_k \)}
        \STATE Sample \( m_k \) points from \( E_k \) to generate set \( \mathcal{S}_k \)
        \STATE Transform \( \mathcal{S}_k \) into high-dimensional representation \( \mathcal{M}_k \) using Mixer Network \( p_\psi \)
        \STATE Process \( \mathcal{M}_k \) and obtain \( \mathcal{O}_k \) using Set Processor \( f_\phi \)
        \STATE Predict solutions \( u_k \) using PDE Probe \( q_\eta \)
    \ENDFOR
    \STATE Compute set-wise physics loss as defined in Equation \ref{eq:setpinns_loss_energy}:
    \[
    \mathcal{L}_{\text{SetPINNs}} = \sum_{X \in \{\Omega, \Omega_0, \partial\Omega\}} \lambda_X \frac{1}{K} \sum_{k=1}^{K} \mathcal{E}_X(E_k, u_\theta)
    \]
    where \( \mathcal{E}_X(E_k, u_\theta) \) is the localized residual energy.
    \STATE Update model parameters \( \psi \), \( \phi \), \( \eta \).
\ENDFOR
\end{algorithmic}
\end{algorithm}

\paragraph{Evaluation.}

The performance of the models was evaluated using the Relative Root Mean Squared Error (rRMSE), which provides a normalized measure of prediction accuracy. The rRMSE, as defined in Equation \ref{eq:rmse}, quantifies the mean squared differences between the predicted and actual values, normalized by the mean squared value of the actual values. This metric ensures comparability across different scales and datasets while effectively capturing deviations in the models' predictions.

\begin{equation}
\label{eq:rmse}
    \text{rRMSE} = \sqrt{\frac{\sum_{n=1}^N (\hat{y}_n - y_n)^2}{\sum_{n=1}^N y_n^2}}
\end{equation}

Our training and evaluation protocols are meticulously designed to ensure robust comparisons and contribute to the reproducibility of the presented results. All choices are made to ensure a fair and comprehensive assessment of the proposed method against strong baselines.

\section{PDE Setup and Their Failure Modes}
\label{app:pde_setup}

For the PDE Setup, we follow the setup of \citet{zhao2024pinnsformer} and \citet{wu2024ropinn}, as it is standard, diverse, and challenging. The selection of PDE benchmarks for our empirical validation is intentionally diverse and challenging. This suite of problems is chosen to rigorously test the capabilities and robustness of SetPINNs against known failure modes, ensuring a thorough assessment across various physical phenomena.

\paragraph{Convection}
The convection equation in one-dimensional space is characterized as a hyperbolic PDE, predominantly utilized for modeling the transport of quantities. It is described using periodic boundary conditions as:

\begin{equation}
\label{eq:convection}
\frac{\partial u}{\partial t} + \beta \frac{\partial u}{\partial x} = 0, \quad \text{for } x \in [0, 2\pi], \, t \in [0, 1]
\end{equation}

\begin{align*}
\text{IC:} & \quad u(x, 0) = \sin(x) \\
\text{BC:} & \quad u(0, t) = u(2\pi, t)
\end{align*}

In this setup, $\beta$ symbolizes the convection coefficient. Here, $\beta = 50$ is chosen to observe the impact on the solution's frequency.

\paragraph{1D Reaction}
The one-dimensional reaction PDE, another hyperbolic PDE, is employed to simulate chemical reaction processes. It employs periodic boundary conditions defined as:

\begin{equation}
\frac{\partial u}{\partial t} - \rho u(1 - u) = 0, \quad \text{for } x \in [0, 2\pi], \, t \in [0, 1]
\end{equation}

\begin{align*}
\text{IC:} & \quad u(x, 0) = \exp\left(- \frac{(x - \pi)^2}{2(\pi/4)^2}\right) \\
\text{BC:} & \quad u(0, t) = u(2\pi, t)
\end{align*}

$\rho$, the reaction coefficient, is set at 5, and the equation's solution is given by:

\begin{equation}
u_{\text{analytical}} = \frac{h(x) \exp(\rho t)}{h(x) \exp(\rho t) + 1 - h(x)}
\end{equation}

where $h(x)$ is based on the initial condition.

\paragraph{1D Wave}
The one-dimensional wave equation, another hyperbolic PDE, is utilized across physics and engineering for describing phenomena such as sound, seismic, and electromagnetic waves. Periodic boundary conditions and system definitions are:

\begin{equation}
\frac{\partial^2 u}{\partial t^2} - \beta \frac{\partial^2 u}{\partial x^2} = 0, \quad \text{for } x \in [0, 1], \, t \in [0, 1]
\end{equation}

\begin{align*}
\text{IC:} & \quad u(x, 0) = \sin(\pi x) + \frac{1}{2} \sin(\beta \pi x), \quad \frac{\partial u(x, 0)}{\partial t} = 0 \\
\text{BC:} & \quad u(0, t) = u(1, t) = 0
\end{align*}

Here, $\beta$, representing the wave speed, is set to 4. The analytical solution is formulated as:

\begin{equation}
u(x, t) = \sin(\pi x) \cos(2\pi t) + \frac{1}{2} \sin(\beta \pi x) \cos(2 \beta \pi t)
\end{equation}

\paragraph{2D Navier-Stokes}
The Navier-Stokes equations in two dimensions describe the flow of incompressible fluids and are a central model in fluid dynamics. These parabolic PDEs are formulated as follows:

\begin{align}
\frac{\partial u}{\partial t} + \lambda_1 \left(u \frac{\partial u}{\partial x} + v \frac{\partial u}{\partial y}\right) &= -\frac{\partial p}{\partial x} + \lambda_2 \left( \frac{\partial^2 u}{\partial x^2} + \frac{\partial^2 u}{\partial y^2} \right) \\
\frac{\partial v}{\partial t} + \lambda_1 \left(u \frac{\partial v}{\partial x} + v \frac{\partial v}{\partial y}\right) &= -\frac{\partial p}{\partial y} + \lambda_2 \left( \frac{\partial^2 v}{\partial x^2} + \frac{\partial^2 v}{\partial y^2} \right)
\end{align}

Here, $u(t, x, y)$ and $v(t, x, y)$ represent the fluid's velocity components in the x and y directions, and $p(t, x, y)$ is the pressure field. For this study, $\lambda_1 = 1$ and $\lambda_2 = 0.01$ are selected. The system does not have an explicit analytical solution, but the simulated solution is provided by \cite{raissi2019physics}.

\paragraph{Clamped Plate}
The clamped plate problem models the steady-state deformation of a thin elastic plate under a localized load. It is governed by the Poisson equation:

\begin{equation}
- \Delta u(x, y) = f(x, y), \quad \text{for } (x, y) \in [0, 1]^2
\end{equation}

with boundary condition \( u(x, y) = 0 \) on \( \partial \Omega \). The forcing term is defined as:

\begin{equation}
f(x, y) =
\begin{cases}
Q, & \text{if } x_0 \le x \le x_1 \text{ and } y_0 \le y \le y_1 \\
0, & \text{otherwise}
\end{cases}
\end{equation}

where \( Q = 20 \), \( x_0 = 0.25 \), \( x_1 = 0.3 \), \( y_0 = 0.7 \), and \( y_1 = 0.75 \).

\paragraph{Harmonic}
This problem models a steady-state response to a smooth sinusoidal forcing. The governing equation is the Poisson equation with a harmonic source:

\begin{equation}
- \Delta u(x, y) = A \sin(k_x \pi x) \sin(k_y \pi y), \quad \text{for } (x, y) \in [0, 1]^2
\end{equation}

subject to Dirichlet boundary conditions \( u(x, y) = 0 \) on \( \partial \Omega \). In our setup, we use \( A = 500 \), \( k_x = 5 \), and \( k_y = 3 \). The exact solution is smooth and periodic within the domain, making this a useful benchmark for assessing spatial resolution and inductive bias.

\paragraph{3D Helmholtz}
The 3D Helmholtz equation models steady-state wave propagation in bounded domains and appears in acoustics, electromagnetics, and quantum mechanics. The governing equation is:

\begin{equation}
- \Delta u(x, y, z) - \kappa^2 u(x, y, z) = f(x, y, z), \quad \text{for } (x, y, z) \in [0, 1]^3
\end{equation}

with Dirichlet boundary condition \( u(x, y, z) = 0 \) on \( \partial \Omega \). We consider a smooth forcing function of the form:

\begin{equation}
f(x, y, z) = A \sin(k_x \pi x) \sin(k_y \pi y) \sin(k_z \pi z)
\end{equation}

where \( A \) is the amplitude and \( k_x, k_y, k_z \) control spatial frequency. The wavenumber is set to \( \kappa = \sqrt{k_x^2 + k_y^2 + k_z^2} \pi \), ensuring the system is consistent with the source term. This setup yields an analytical solution of the same sinusoidal form, providing a controlled benchmark for evaluating 3D modeling accuracy.

\subsection{Failure Modes in PDEs}
Table~\ref{tab:pde_failure_modes} summarizes the key modeling challenges associated with each PDE benchmark used in our evaluation and outlines common failure modes encountered by standard PINNs. These include difficulties such as resolving sharp gradients, handling high-frequency components, satisfying boundary conditions, and maintaining stability in complex multi-scale systems. The diversity of these challenges highlights the need for architectures like SetPINNs that are better equipped to handle local structure and mitigate known weaknesses of conventional PINNs.

\begin{table}[h]
\centering
\caption{PDE Benchmark Challenges and Typical PINN Failure Modes. This table outlines modeling challenges for each benchmark and common difficulties for standard PINNs.}
\label{tab:pde_failure_modes}
\begin{tabular}{p{2.5cm} p{4.5cm} p{4.5cm}} 
\toprule
\textbf{PDE Benchmark} & \textbf{Primary Modeling Challenge(s)} & \textbf{Typical PINN Failure Modes / Difficulties} \\
\midrule
1D Reaction-Diffusion & Sharp moving fronts; steep gradients & Poor gradient resolution; over-smoothing; numerical stiffness; slow convergence. \\
\midrule
1D Wave Equation & Sharp wavefront propagation; high-frequency components; dispersion/dissipation risks & Difficulty with high-frequencies; poor sharp feature propagation; low-frequency spectral bias. \\
\midrule
Convection Equation & Sharp profile transport; numerical diffusion/oscillations (Gibbs) & False diffusion smoothing profiles; directional bias; oscillations near discontinuities. \\
\midrule
2D Navier-Stokes & Coupled nonlinearities (velocity-pressure); multi-scale flow features (vortices, boundary layers) & Difficulty resolving strong nonlinearities \& fine-scale capture; instability; loss balancing. \\
\midrule
Harmonic Poisson & Boundary condition (BC) sensitivity; solution smoothness (Laplacian) & Imprecise BC satisfaction (esp. complex); slow convergence for large domains. \\
\midrule
Clamped Plate (Localized Load) & Localized stress/strain; sharp features at load/boundaries & Misses peak stress (poor local resolution/supervision); clamped boundary inaccuracies. \\
\midrule
3D Helmholtz & High-dimensionality; oscillatory solutions (eigenfunctions); pollution error (error grows with freq./wavenumber) & Poor 3D high-frequency resolution; high computational cost; severe spectral bias; pollution error (inaccurate phase/amplitude). \\
\bottomrule
\end{tabular}
\end{table}

\section{Computational Complexity and Scalability.}
\label{app:computation_complexity}
SetPINNs achieve improved modeling performance without incurring significant computational overhead. Table~\ref{tab:complexity_comparison} compares the theoretical complexity of SetPINNs with standard PINNs and PINNsFormer. While standard PINNs have linear complexity $\mathcal{O}(M)$ in the number of collocation points \(M\), PINNsFormer incurs a higher cost of $\mathcal{O}(K^2 \cdot M)$ due to attention over pseudo-sequences of size \(K\). In contrast, SetPINNs process sets of size \(K\), resulting in a computational cost of $\mathcal{O}(K^2 \cdot \frac{M}{K}) = \mathcal{O}(K \cdot M)$, which is practically efficient and scales favorably.

Moreover, unlike traditional Finite Element Methods (FEMs) that require repeated meshing and basis construction, SetPINNs perform domain partitioning and sampling only once at initialization. As demonstrated in our runtime analysis (Appendix~\ref{app:compute}), the training time of SetPINNs is comparable to baselines despite their structural advantages. Crucially, the computational cost of SetPINNs is independent of the underlying domain or PDE dimensionality after partitioning, making them especially attractive for high-dimensional problems.

\begin{table}[h]
\centering
\caption{Computational complexity of SetPINNs compared to baselines. \(M\): total collocation points; \(K\): set size or pseudo-sequence length.}
\label{tab:complexity_comparison}
\begin{tabular}{lcc}
\toprule
\textbf{Method} & \textbf{Complexity} & \textbf{Remark} \\
\midrule
PINNs & $\mathcal{O}(M)$ & Pointwise MLPs over all $M$ collocation points \\
SetPINNs & $\mathcal{O}\left(K^2 \cdot \frac{M}{K} \right)$ & $M$: collocation points. \(K\): set-size. \\
PINNsFormer & $\mathcal{O}(K^2 \cdot M)$ &  $M$: collocation points. \(K\): set-size. \\
\bottomrule
\end{tabular}
\end{table}

These properties make SetPINNs highly scalable while preserving their ability to model local dependencies. This is particularly beneficial for problems in high dimensions or with large spatial domains, where conventional PINNs may become infeasible due to computational bottlenecks.

This complexity analysis underscores that SetPINNs' enhanced modeling power is achieved without undue computational burden. This favorable balance highlights its practical scalability for complex, potentially high-dimensional, scientific and engineering applications, marking a significant step towards efficient physics-informed learning.

\section{Experimental Details}
\label{app:experiments}

\subsection{Hyperparameters}
\label{app:hyperparameters}
Full transparency in experimental configuration is paramount for scientific validation and reproducibility. We therefore provide a comprehensive account of all model hyperparameters and specific training settings used in our empirical evaluations, facilitating verification by the research community.

\paragraph{Model hyperparameters.}
Table \ref{tab:model-hyperparameters} outlines the hyperparameters for the different models evaluated in this study, including Physics-Informed Neural Networks (PINNs), Quadratic Residual Networks (QRes), First-Layer Sine (FLS), PINNsFormer, and SetPINNs. Each model is configured with a specific number of hidden layers and hidden sizes. The PINNsFormer and SetPINNs models also include additional parameters such as the number of encoders and decoders, embedding size, and the number of attention heads. These configurations are crucial for defining the model architectures and their capacities to learn from the data. Table \ref{tab:total-params} shows the total parameters of all models. For a fair comparison, all models have relatively similar numbers of trainable parameters. For implementation, we follow the same implementation pipeline as of PINNsFormer\footnote{https://github.com/AdityaLab/pinnsformer} and use their implementation of PINNs, QRes, FLS, and PINNsFormer. For fair comparisons, the model architecture of SetPINNs is kept consistent with PINNsFormer.

\begin{table}[h]
    \centering
    \caption{Hyperparameters for the different models evaluated in the study, including the number of hidden layers, hidden sizes, and additional parameters for complex models like PINNsFormer and SetPINNs.}
    \label{tab:model-hyperparameters}
    \begin{tabular}{c | c | c}
        \hline \hline
         Model & Hyperparameter & Value \\
         \hline
         \multirow{2}{*}{PINNs} & hidden layers & 4  \\
         &hidden size & 512\\
         \hline
         \multirow{2}{*}{QRes} & hidden layers & 4  \\
         &hidden size & 256\\
         \hline
         \multirow{2}{*}{FLS} & hidden layers & 4  \\
         &hidden size & 512\\
         \hline
         \multirow{7}{*}{PINNsFormer} & $k$ & 5  \\
         &$\Delta t$ & 1e-4\\
         & \# of encoder & 1\\
         & \# of decoder& 1\\
         &embedding size& 32\\
         &head& 2\\
         &hidden size& 512\\
         \hline
         \multirow{2}{*}{RoPINNs} & hidden layers & 4  \\
         &hidden size & 512\\
         \hline
         \multirow{6}{*}{SetPINNs} & set size & 4 \\
         & \# of encoder & 1\\
         &embedding size& 32\\
         &head& 2\\
         &hidden size& 512\\
        \hline \hline
    \end{tabular}
\end{table}

\begin{table}[h]
    \centering
    \caption{Total number of trainable parameters for all models. For a fair comparison, all models have relatively similar numbers of trainable parameters.}
    \label{tab:total-params}
    \begin{tabular}{c | c }
        \hline \hline
         Model & Total trainable parameters \\
        \hline
         PINNs & 527K \\
         FLS & 527K \\
         QRes & 397K \\
         PINNsFormer & 454K \\
         RoPINNs & 527K \\
         SetPINNs & 366K \\
        \hline \hline
    \end{tabular}
\end{table}

\paragraph{Training Hyperparameters.}
The training process for the models utilized specific hyperparameters as listed in Table \ref{tab:training-hyperparameters}. The optimization followed a two-stage approach, first using the Adam optimizer for a set number of iterations, followed by fine-tuning with the L-BFGS optimizer. The L-BFGS optimization employed the strong Wolfe line search condition. The weighting parameters \( \lambda_{\Omega} \), \( \lambda_{\Omega_0} \), and \( \lambda_{\partial\Omega} \) were set to specific values to balance the contributions of the PDE residual, initial conditions, and boundary conditions in the loss function (Equation \ref{eq:setpinns_loss_energy}). These parameters were kept consistent across all models to ensure a fair comparison and reliable convergence.

\begin{table}[h]
    \centering
    \caption{Training hyperparameters used for all models, including optimizer settings, weighting parameters \( \lambda_{\Omega} \), \( \lambda_{\Omega_0} \), and \( \lambda_{\partial\Omega} \), and dataset splits. Missing values are placeholders that should be updated based on the main paper.}
    \label{tab:training-hyperparameters}
    \begin{tabular}{c | c }
        \hline \hline
         Hyperparameter & Value \\
        \hline
         Adam Iterations & 100 \\
         L-BFGS Iterations & 2000 \\
         L-BFGS Line Search Condition & Strong Wolfe \\
         $\lambda_{\Omega}$ & 1 \\
         $\lambda_{\Omega_0}$ & 1 \\
         $\lambda_{\partial\Omega}$ & 1 \\
         Train:Test Split (grid) & $50\times50$:$101\times101$ \\
         Train:Test Split (number of samples) & 2500:10201\\
        \hline \hline
    \end{tabular}
\end{table}

\subsection{Runtime Analysis}
\label{app:compute}

All models are implemented in PyTorch and trained separately on a single NVIDIA A100 GPU. The runtime of different methods is evaluated on the convection equation, averaged over ten independent runs. The results, reported in seconds, are summarized in Table~\ref{tab:runtime}.

\begin{table}[h]
    \centering
    \caption{Mean and standard deviation of runtime (in seconds)/wall-clock timings for different models, averaged over ten runs on the convection equation. High variance is primarily due to the L-BFGS optimizer, which slows down significantly when encountering challenging optimization landscapes with local minima. The proposed SetPINNs model achieves a lower mean runtime and variance compared to RoPINNs and PINNsFormer, demonstrating more stable and efficient training.}
    \label{tab:runtime}
    \begin{tabular}{c | c}
        \hline \hline
        Model & Runtime (Mean ± Std) [s] \\
        \hline
        PINNs & 314.43 ± 6397.41 \\
        QRes & 266.95 ± 2155.99 \\
        FLS & 369.03 ± 5217.46 \\
        RoPINNs & 619.04 ± 7665.12 \\
        PINNsFormer & 1015.64 ± 13389.03 \\
        SetPINNs (Ours) & 614.99 ± 1540.60 \\
        \hline \hline
    \end{tabular}
\end{table}

The high variance in runtime across models is primarily due to the L-BFGS optimizer, which exhibits instability when encountering challenging optimization landscapes with local minima. This leads to varying optimization times across different runs. 

Among the state-of-the-art methods, RoPINNs and PINNsFormer demonstrate particularly high variance, indicating sensitivity to training conditions. In contrast, the proposed SetPINNs model achieves a more stable and efficient training process, with lower mean runtime and variance compared to RoPINNs and PINNsFormer.

\paragraph{GPU Memory.} 
The GPU memory footprint remains similar across models, as the number of trainable parameters is kept approximately the same for fair comparisons. All models are trained on a single-node NVIDIA A100 GPU.

\subsection{Reproducibility.} 
To ensure transparency and reproducibility, we provide all hyperparameters, training details, and implementation choices in the manuscript, including Appendix~\ref{app:hyperparameters}. Additionally, code is made available in the supplementary. It is important to note that some of the baseline results reported in our study may not exactly match the values originally published in their respective works. This discrepancy arises because we conduct multiple independent runs and report the mean and variance to provide a more comprehensive evaluation of model performance and robustness. By doing so, we ensure a fair comparison and account for variability in training due to factors such as optimizer dynamics and stochastic sampling.

\section{Real-world Process Engineering Tasks}
\subsection{Predicting Activity Coefficients}
\label{app:activity}
Activity coefficients are a central thermodynamic property describing the behavior of components in mixtures. In determining the deviation from the ideal mixture, activity coefficients are fundamental for modeling and simulating reaction and separation processes, such as distillation, absorption, and liquid-liquid extraction. In a mixture, each component has an individual activity coefficient that highly depends on the molecular structure of the components as well as on the temperature and concentration in the mixture (the latter is typically given in mole fractions of the components ranging between 0 and 1). 

Since measuring activity coefficients is exceptionally time-consuming and expensive, experimental data on this property is scarce. Therefore, prediction methods are established in practice; the most commonly used ones are based on physical theories. Compared to machine-learning models for predicting activity coefficients, which have also been proposed in the literature \citep{jirasek2020machine}, the most significant advantage of physical models is that they comply with thermodynamic consistency criteria, such as the Gibbs-Duhem equation that relates the activity coefficients within a mixture. For binary mixtures composed of two components and the isothermal and isobaric case, the Gibbs-Duhem equation reads: 
\begin{equation}
     x_1\left (\frac{\partial \text{ln}\gamma_{1} }{\partial x_1}  \right )_{T}+ (1-x_1)\left (\frac{\partial \text{ln}\gamma_{2} }{\partial x_1}  \right )_{T}=0
    \label{Gibbs-Duhem}
\end{equation}
, where $\text{ln}\gamma_{1}$ and $\text{ln}\gamma_{2} $ are the logarithmic activity coefficients of the two components that make up the mixture, $T$ is the temperature, and $x_1$ is the mole fraction of the first component.


\paragraph{Setup.}
The experimental activity coefficient data set was taken from the Dortmund Data Bank (DDB). Data on activity coefficients at infinite dilution were directly adopted from the DDB. Furthermore, activity coefficients were calculated from vapor-liquid equilibrium data from the DDB. In the preprocessing, data points labeled as poor quality by the DDB were excluded. Furthermore, only components for which a SMILES (simplified molecular-input line-entry system) string could be retrieved using the CAS number (preferred) or the component name using the Cactus database were considered. SMILES are a chemical language used to describe the molecular structure of a molecule. These SMILES were then converted into canonical SMILES using RDKit, which resulted in dropping a few SMILES that could not be converted. 
One system is defined as the combination of two components. The train-test split was done in a system-wise manner, whereby 10 \% of the systems were used for the test set. RDKit was used to create molecular descriptors for each component. Specifically, a count-based Morgan fingerprint with zero radius and bitesize of 128 was used.

\subsection{Predicting Agglomerate Breakage}
\label{app:agglomerate}
Population balance equations (PBE) are a well-known and general method for calculating the temporal evolution of particle property distributions $u(x,t)$, with ever-increasing applications in a variety of fields \cite{Ramkrishna2014}. 
In general, PBE are integro-differential equations and require numerical solutions. We investigate the one-dimensional case of pure agglomerate breakage, where analytical solutions exist for specific boundary conditions. The PBE is given by:

\begin{equation}
\frac{\partial u(x,t)}{\partial t}= \int\limits_{x}^{\infty}f(x,x')r(x')u(x',t)\mathrm{d}x'-r(x)u(x,t) , \label{eq:popdis} 
\end{equation}%

\begin{align*}
\text{IC:} & \quad u(x, 0) = \delta(x-L) \\
\text{BC:} & \quad r(x)=x \\ 
           & \quad f(x,y)=\frac{2}{y}
\end{align*}


Here, the breakage rate $r$ and breakage function $f$ are the so-called kernels and define the physical behavior of the system. The analytical solution for this special case is formulated as \cite{RN772}:

\begin{equation}
    u(x,t) = \mathrm{exp}\left(-tx\right)\left(\delta(x-L)+\left[2t+t^2(L-x)\right]\theta(L-x)\right) 
\end{equation}

with $\delta$ being dirac delta and 

\begin{equation}
    \theta(x-L)=
    \begin{cases}
        1, & x<L \\
        0, & \mathrm{otherwise}
    \end{cases}
    \quad .
\end{equation}

In a real-world setting, i.e. when only experimental data is available, the kernel values are generally unknown. Although empirical equations exist, they have to be calibrated to experiments. This makes benchmarking solely on experimental data impossible and hence, we used the provided special case. However, it should be emphasized that PINNs have already been applied for solving the inverse problem, i.e. estimating unknown kernels \cite{RN723}. Therefore, an improved accuracy on synthetic data will likely correspond to higher accuracy of the inverse problem, when applied to experimental data.




\subsection{Detailed Discussion of Chemical Process Engineering Results}
\label{app:chemical_engineering_discussion}

SetPINNs excel in chemical engineering tasks like predicting activity coefficients (AC) and modeling agglomerate breakage (AB) by effectively addressing the unique nature of real-world scientific data and system symmetries.

\begin{enumerate}
    \item \textbf{Handling Irregular, Sparse, and Variable-Sized Data:}
    Chemical data (e.g., AC from DDB \citep{DDB2023}) are often sparse, non-uniform, and involve systems with varying numbers of components (e.g., binary, ternary mixtures). Standard PINNs, requiring fixed-size, ordered inputs, struggle with such data. SetPINNs natively process sets of arbitrary cardinality, making them robust to data irregularity and directly applicable to systems of varying sizes without ad-hoc modifications.

    \item \textbf{Exploiting Permutation Invariance/Equivariance:}
    Physical properties of mixtures or particulate systems are often invariant to component order. SetPINNs' permutation-equivariant architecture (via attention) aligns with this fundamental physical inductive bias. This allows the model to learn true underlying relationships without being confounded by artificial input orderings, unlike standard PINNs which impose such order.

    \item \textbf{Superior Performance with Physical Consistency:}
    SetPINNs achieved state-of-the-art accuracy on both AC (rRMSE $0.090 \pm 0.00$) and AB tasks (Table~\ref{tab:overall}), outperforming even established physics-based models like UNIFAC (AC rRMSE $0.166$) while simultaneously enforcing physical constraints (e.g., Gibbs-Duhem for AC). This suggests SetPINNs capture subtle interaction patterns from sparse data beyond traditional assumptions, yet remain physically consistent.

    \item \textbf{Natural Representation for Interaction Learning:}
    Representing chemical systems as sets is physically natural. The attention mechanism in SetPINNs efficiently learns pairwise or higher-order interactions crucial for emergent system properties (e.g., non-ideal mixing, collision-induced breakage). This set-based approach, combined with physics-informed learning, promotes generalization and scalability, bridging data-driven and physics-based modeling.
\end{enumerate}

In essence, SetPINNs' ability to process set-structured, permutation-invariant data while incorporating physical laws makes them fundamentally well-suited for a significant class of scientific problems, leading to superior accuracy, physical consistency, and better generalization in complex, real-world chemical engineering systems.

The successful application of SetPINNs to these complex, real-world chemical engineering tasks, which inherently involve permutation-invariant systems and sparse data, robustly demonstrates the method's practical utility and its potential to address significant challenges in specialized scientific domains by effectively integrating data-driven insights with fundamental physical principles.

\begin{figure}
    \centering
    \includegraphics[scale=0.9]{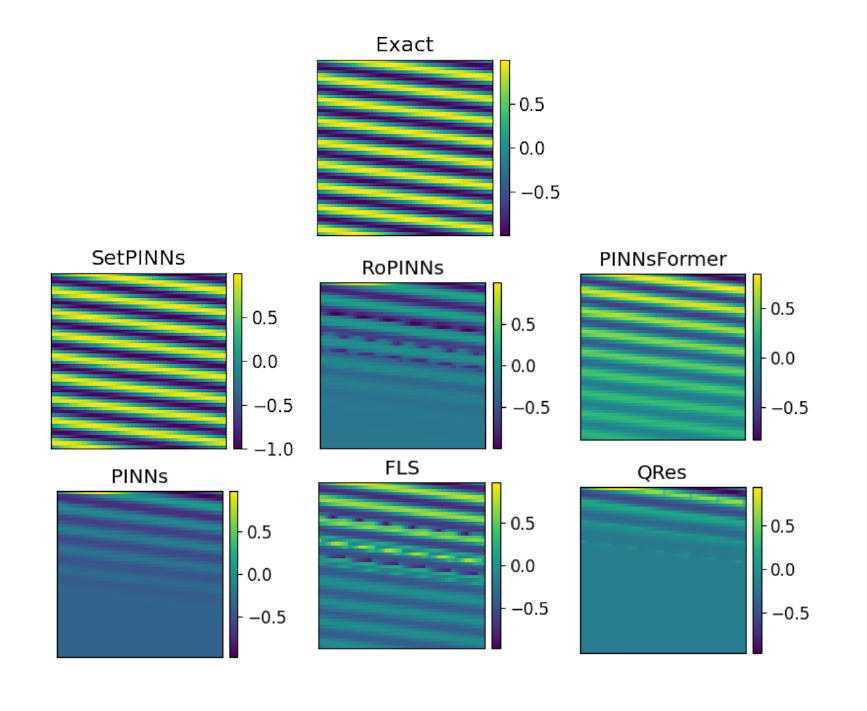}
    \caption{Qualitative comparison of predicted solutions for the convection equation (Eq~\ref{eq:convection}) across different PINN variants. The top panel shows the ground truth (“Exact”) solution. SetPINNs closely match the true solution. Other baselines—RoPINNs, PINNsFormer, FLS, QRes, and standard PINNs—exhibit varying degrees of degradation. This highlights the advantage of SetPINNs in capturing fine-grained local patterns.}
    \label{fig:convection}
\end{figure}

\section{Ablation Studies}
\label{app:ablation}
To further elucidate the contributions of individual components and the sensitivity to key design choices within the SetPINN framework, we conduct a series of thorough ablation studies. These investigations provide deeper insights into the model's behavior and validate its architectural integrity.

We examine the effects of different element sizes, the number of attention heads, and the number of block in the encoder, and apply element-aware sampling on PINNs to observe changes in error rates. For these ablation studies, we select a 1D reaction equation and use 100 residual, boundary, and initial points. The domain is discretized with a 100x100 grid, which is kept consistent across all experiments. All other hyperparameters remain the same unless specifically mentioned. We report rRMSE across all the experiments in ablation studies.

\paragraph{Qualitative Evaluatio on Convection Equation}

Figure~\ref{fig:convection} presents a qualitative comparison of predicted solutions for the convection equation (Eq.~\ref{eq:convection}) across various PINN variants. SetPINNs closely align with the ground truth, accurately capturing sharp fronts and fine-grained local structures. In contrast, baselines such as RoPINNs, PINNsFormer, FLS, QRes, and standard PINNs display visible artifacts or smoothing, indicating difficulty in modeling localized transport behavior. These results underscore the effectiveness of SetPINNs in preserving high-frequency features critical to convection dynamics.

\paragraph{Effect of EAS on Conventional PINNs.}

Table~\ref{tab:ews-effect} presents the relative rRMSE (mean $\pm$ standard deviation) of vanilla PINNs under different sampling strategies across five benchmark PDEs. In addition to uniform random sampling, we evaluate Latin Hypercube Sampling (LHS), Residual-based Adaptive Distribution (RAD) \citep{wu2023comprehensive}, and our proposed Element-Aware Sampling (EAS). LHS provides stratified coverage of the input domain, while RAD adaptively refines the sampling distribution by focusing on high-residual regions. Among these, EAS generally yields the lowest error, indicating its effectiveness in improving supervision without architectural changes.

\begin{table}[ht]
    \centering
    \caption{Impact of Sampling Strategies on Vanilla PINN Performance (rRMSE $\pm$ Std. Dev.). Element-Aware Sampling (EAS) generally improves accuracy.}
    \label{tab:ews-effect}
    \begin{tabular}{l c c c c}
    \toprule
    \textbf{PDE} & \textbf{Vanilla} & \textbf{+ EAS} & \textbf{+ LHS} & \textbf{+ RAD} \\
    \midrule
    1D Wave     & $0.148 \pm 0.01$ & $0.119 \pm 0.02$ & $0.135 \pm 0.02$ & $0.132 \pm 0.03$ \\
    1D Reaction & $0.801 \pm 0.12$ & $0.725 \pm 0.16$ & $0.766 \pm 0.15$ & $0.752 \pm 0.14$ \\
    Convection  & $1.136 \pm 0.13$ & $0.905 \pm 0.07$ & $1.020 \pm 0.10$ & $0.985 \pm 0.09$ \\
    Harmonic    & $0.342 \pm 0.15$ & $0.270 \pm 0.04$ & $0.295 \pm 0.06$ & $0.310 \pm 0.08$ \\
    Plate       & $1.467 \pm 0.58$ & $1.275 \pm 0.31$ & $1.380 \pm 0.34$ & $1.198 \pm 0.29$ \\
    \bottomrule
    \end{tabular}
\end{table}

\paragraph{Element Size}
\label{app:element_size}
Domain partitioning is a core component of the SetPINNs framework. In this experiment, we study how varying the size of these elements affects prediction accuracy, using the reaction-diffusion equation as a test case. The input domain is discretized into a $100 \times 100$ grid and further partitioned into square elements of size $n \times n$. Each element forms a point set that serves as the input to the set encoder, as defined by the architecture hyperparameters in Table~\ref{tab:model-hyperparameters}. For instance, a $2 \times 2$ element corresponds to a set size of 4, while a $4 \times 4$ element yields a set size of 16, and so on. Although we restrict our analysis to square elements here, the method generalizes to arbitrary shapes.

\begin{figure}[h]
    \centering
    \includegraphics[scale=0.8]{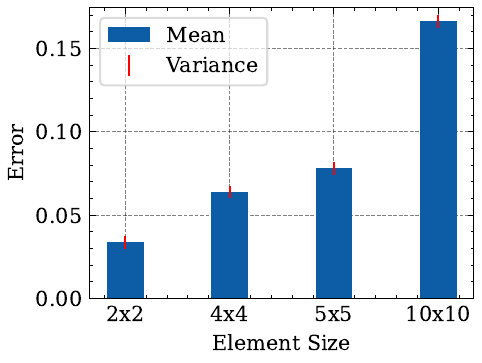}
    \caption{Bar chart illustrating the impact of element size on error rates. As shown, larger element sizes result in significantly higher error rates, which is attributed to the increased distance between points within the sets for larger elements. This trend empirically validates our theoretical analysis.}
    \label{fig:element_ablation}
\end{figure}

The results, shown in Figure~\ref{fig:element_ablation}, reveal a clear trend: prediction error increases with element size. This degradation is attributed to the growing spatial distance between points within each set. Smaller elements preserve local context, allowing the model to exploit short-range correlations more effectively. In contrast, larger elements span wider areas, weakening intra-set interactions and reducing the model’s ability to capture local dependencies. These empirical findings are consistent with our theoretical analysis, which predicts diminished performance when points attending to each other lie farther apart. This experiment thus validates the importance of maintaining locality in the design of element-wise set structures.

\paragraph{No of Transformer blocks}
In this experiment, we explore the effect of varying the number of Transformer blocks on error rates. We conduct experiments using SetPINNs with hyperparameters specified in Table~\ref{tab:model-hyperparameters}, testing configurations with both 1 and 2 Transformer blocks (N) in the encoder.

\begin{figure}[h]
    \centering
    \includegraphics[scale=0.8]{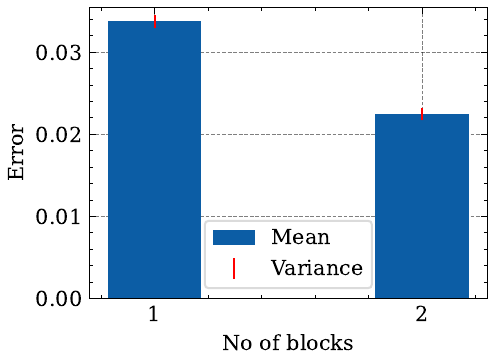}
    \caption{Bar chart illustrating the impact of the number of transformer blocks on error rates. The x-axis represents the number of blocks in the encoder, and the y-axis shows the corresponding error rates. The results indicate that using 2 blocks leads to significantly better performance compared to using just 1 block, likely due to the increased capacity for learning complex representations.}
    \label{fig:bar_N}
\end{figure}

The results, depicted in Figure~\ref{fig:bar_N}, show that using 2 transformer blocks significantly improves the results compared to using just 1 block. This improvement can be attributed to the additional layers providing more capacity for the model to learn complex representations, thereby enhancing performance.

\paragraph{No of attention heads}
In this experiment, we investigate the impact of varying no of attention heads on error rates. For this purpose, we utilize SetPINNs with hyperparameters outlined in Table~\ref{tab:model-hyperparameters}, employing an embedding size of 256 instead of 32 to accommodate a greater number of heads.

\begin{figure}[h]
    \centering
    \includegraphics[scale=0.8]{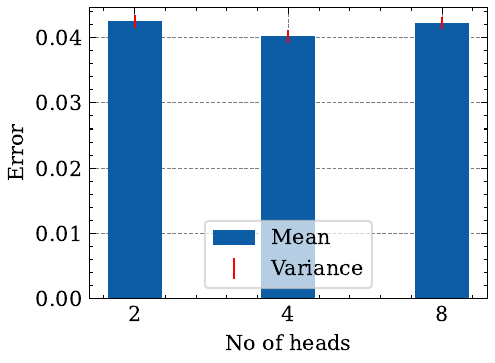}
    \caption{Bar chart illustrating the impact of varying the number of attention heads on error rates. The x-axis represents the number of attention heads, and the y-axis shows the corresponding error rates. The results indicate that changing the number of attention heads does not significantly affect the error rates, which remain consistent across different experiments. This suggests that the additional attention heads do not contribute to improved performance, possibly due to the model already capturing the necessary information with fewer heads.}
    \label{fig:heads_ablation}
\end{figure}

The results, shown in Figure~\ref{fig:heads_ablation}, reveal that the number of attention heads does not significantly impact error rates, as the error rates remain consistent across all experiments.rates remain similar across all the experiments. This could be because the increased number of attention heads does not provide additional useful representational capacity or because the model already captures the necessary information with fewer heads, leading to diminishing returns with more heads.

\section{Synopsis of Key Contributions}
This paper introduced SetPINNs, a novel and impactful framework that significantly advances physics-informed learning. By uniquely integrating element-aware sampling with expressive set-based neural networks, SetPINNs masterfully model local dependencies, leading to substantial gains in solution accuracy and training stability. Our rigorous theoretical analysis provides a solid foundation, demonstrating lower-variance estimates and enhanced domain coverage. These critical advantages are consistently validated across extensive experiments encompassing both synthetic and large real-world benchmarks, including complex and high-dimensional PDEs. SetPINNs consistently demonstrate superior performance against a comprehensive suite of recent and highly relevant state-of-the-art baselines. Furthermore, our scalability analysis and wall-clock runtime comparisons clearly favor SetPINNs, highlighting its practical efficiency. SetPINNs thus offer a robust, highly effective, and well-grounded method, marking a clear and significant step forward in leveraging deep learning for complex scientific simulations and promising broad, impactful applicability.

\section{Broader Impact}
\label{app:broader_impact}

Our work on SetPINNs contributes to advancing the field of machine learning by improving the accuracy and stability of physics-informed neural networks (PINNs) for solving partial differential equations (PDEs). This has broad implications across scientific computing, engineering, and industrial applications, where PDEs play a fundamental role in modeling complex physical phenomena.

A key societal benefit of SetPINNs is their potential to enhance computational efficiency and reliability in fields such as climate modeling, fluid dynamics, material science, and structural engineering. By improving the accuracy of physics-driven machine learning models, our approach can lead to more precise simulations, reducing reliance on costly and resource-intensive physical experiments. This could positively impact sustainability efforts by optimizing energy consumption in simulations and minimizing waste in experimental studies.

However, as with any machine learning model, there are considerations regarding fairness and responsible deployment. SetPINNs rely on training data and numerical solvers, and their effectiveness depends on the quality of these inputs. In critical applications such as engineering safety assessments or environmental forecasting, improper usage or over-reliance on imperfect models could lead to incorrect predictions with real-world consequences. It is essential that users validate SetPINNs’ predictions against ground truth data and established numerical methods when applied to high-stakes domains.

We do not foresee any direct ethical concerns or harmful societal implications arising from this research. Instead, SetPINNs serve as a step toward more robust, efficient, and generalizable machine learning methods for solving PDEs, benefiting a wide range of scientific and industrial applications.

We are optimistic that SetPINNs, developed with a strong consideration for responsible and ethical application, will serve as a valuable and impactful tool for advancing scientific discovery and engineering innovation across a multitude of disciplines. The framework's design promotes both accuracy and interpretability, contributing positively to the ongoing integration of AI in scientific research.

\end{document}